\documentclass[a4paper,10pt]{article}

\pdfoutput=1
\usepackage{fullpage}
\usepackage[utf8]{inputenc}
\usepackage{graphicx}
\usepackage{amssymb}
\usepackage{amsthm}
\usepackage{amsmath}
\usepackage{amsfonts}
\usepackage{tabularx,colortbl}
\usepackage{color}
\usepackage[english]{babel}
\usepackage[pdftex]{hyperref}
\usepackage{subfigure}
\usepackage{algorithm}
\usepackage{algorithmic}
\usepackage[numbers,sort&compress]{natbib}
\usepackage{authblk}
\usepackage{appendix}

\newtheorem{theorems}{theorems}
\newtheorem{thmThetaMST}[theorems]{Theorem}
\newtheorem{thmSequenceOrdering}[theorems]{Theorem}
\newtheorem{thmTheta}[theorems]{Theorem}

\definecolor{Orange}{rgb}{1,0.64,0}
\definecolor{lgray}{rgb}{0.9,0.9,0.9}

\newcommand{\argmax}{\operatornamewithlimits{arg\ max}}
\newcommand{\argmin}{\operatornamewithlimits{arg\ min}}

\title{Designing labeled graph classifiers by exploiting the {R}\'{e}nyi entropy of the dissimilarity representation}

\author[1]{Lorenzo Livi\thanks{l.livi@exeter.ac.uk}\thanks{Corresponding author}}
\affil[1]{Department of Computer Science, College of Engineering, Mathematics and Physical Sciences, University of Exeter, Exeter EX4 4QF, UK}

\providecommand{\keywords}[1]{\textbf{\textit{Keywords---}} #1}

\begin{document}

\date{}
\maketitle

\begin{abstract}
Representing patterns as labeled graphs is becoming increasingly common in the broad field of computational intelligence.
Accordingly, a wide repertoire of pattern recognition tools, such as classifiers and knowledge discovery procedures, are nowadays available and tested for various datasets of labeled graphs.
However, the design of effective learning procedures operating in the space of labeled graphs is still a challenging problem, especially from the computational complexity viewpoint.
In this paper, we present a major improvement of a general-purpose classifier for graphs, which is conceived on an interplay between dissimilarity representation, clustering, information-theoretic techniques, and evolutionary optimization algorithms. The improvement focuses on a specific key subroutine devised to compress the input data.
We prove different theorems which are fundamental to the setting of the parameters controlling such a compression operation. We demonstrate the effectiveness of the resulting classifier by benchmarking the developed variants on well-known datasets of labeled graphs, considering as distinct performance indicators the classification accuracy, computing time, and parsimony in terms of structural complexity of the synthesized classification models.
The results show state-of-the-art standards in terms of test set accuracy and a considerable speed-up for what concerns the computing time.\\
\end{abstract}

\section{Introduction}

A graph offers a powerful model for representing patterns characterized by interacting elements, in 	both static or dynamic scenarios.
A labeled graph (also called attributed graph) is a tuple $G=(\mathcal{V}, \mathcal{E}, \mu, \nu)$, where $\mathcal{V}$ is the finite set of vertices, $\mathcal{E}\subseteq \mathcal{V}\times \mathcal{V}$ is the set of edges, $\mu: \mathcal{V} \rightarrow \mathcal{L}_{\mathcal{V}}$ is the vertex labeling function, with $\mathcal{L}_{\mathcal{V}}$ denoting the set of vertex labels, and finally $\nu: \mathcal{E} \rightarrow \mathcal{L}_{\mathcal{E}}$ is the edge labeling function, with $\mathcal{L}_{\mathcal{E}}$ denoting the set of edge labels \cite{gm_survey}.
The topology of a graph enables the characterization of a pattern in terms of ``interacting'' elements. Moreover, the generality of both $\mathcal{L}_{\mathcal{V}}$ and $\mathcal{L}_{\mathcal{E}}$ allows to cover a broad range of real-world patterns.
Applications involving labeled graphs for representing data can be cited in many scientific fields, such as electrical circuits \cite{dorfler+bullo2011}, networks of dynamical systems \cite{porfiri2008synchronization}, biochemical networks \cite{ecoli_graph,doi:10.1021/cr3002356}, time-varying labeled graphs \cite{dyngraph_ijcnn_2013}, and segmented images \cite{Serratosa:2013:CRB:2435460.2435704,Noma20121159}.
Owing to the rapid diffusion of (cheap) multicore computing hardware, and motivated by the increasing availability of interesting datasets describing complex interaction-oriented patterns, recent researches on graph-based pattern recognition systems have produced numerous methods \cite{chen2014type,jothi2015hybrid,odse,gralg_2012,marfil_escolano__2009,Bengoetxea20022867,CesarJr20052099,Lozano2006539,Bai_Hancock__2012,eocc,
it_diss_graphs__Escolano_2013,10.1109/TPAMI.2012.226,Shervashidze__2011,Lozano2013177,bai2014quantum,Bunke2012811,emmert2016fifty}.

Focusing on the high-level design of classification systems for graphs, it is possible to identify two main different approaches: those that operate directly in the domain of labeled graphs and those that deal with the classification problem in a suitable embedding space.
Of notable interest are those systems that are based on the so-called \textit{explicit graph embedding} algorithms, which transform the input graphs into numeric vectors by means of a mapping or feature extraction technique \cite{gm_survey}.
Graph embedding algorithms \cite{Gibert20123072,riesen+bunke2010,gralg_2012,robleskelly07Riemannianapproachtographembedding,escolano_2011,Luqman:2013:FMG:2381462.2381562,ZareBorzeshi20131648,5648360}, operate by explicitly developing an embedding space, $\mathcal{D}$.
The distance between two graphs is hence computed processing their vector representations in $\mathcal{D}$, usually by either a geometric or information-theoretic interpretation (e.g., based on divergences \cite{it_diss_graphs__Escolano_2013}).
We distinguish two main categories of graph embedding algorithms: those that are defined in terms of a core inexact graph matching (IGM) procedure working directly in the graph domain, $\mathcal{G}$, and those that exploit a matrix representation of the graph to extract characterizing information.
The former (e.g., see \cite{gralg_2012,riesen+bunke2010,ZareBorzeshi20131648}) can process virtually any type of labeled graph, according to the capability of the adopted core matching algorithm.
The latter \cite{seriation+gradis_lncs_2012,5648360,robleskelly07Riemannianapproachtographembedding,escolano_2011,jain2016geometry,jain2016statistical} are constrained to process a restricted variety of labeled graphs, in which all the relevant information can be effectively encoded into a matrix representation of a graph, such as (weighted) adjacency, transition, or Laplacian matrix.
The interested reader is referred to \cite{gm_survey,Bunke2012811,Hancock2012833} and references therein for reviews of recent graph embedding techniques.

The dissimilarity representation offers a valuable framework for this purpose, since it permits to describe arbitrarily complex objects by means of their pairwise dissimilarity values (DV) \cite{pkekalska+duin2005}.
In the dissimilarity representation, the elements of an input dataset $\mathcal{S}\subset\mathcal{X}$ are characterized by considering vectors made of their pairwise DVs \cite{pkekalska+duin2005,schleif2015indefinite}.
The key component is hence the definition of a nonnegative (bounded) dissimilarity measure $d: \mathcal{X}\times\mathcal{X}\rightarrow\mathbb{R}^{+}$.
A set of prototypes, $\mathcal{R}$, called representation set (RS), is used to develop the dissimilarity matrix (DM), $\mathbf{D}$, whose elements are given as $D_{ij}=d(x_i, r_j)$, for every $x_i\in\mathcal{S}$ and $r_j\in\mathcal{R}$.
By means of $\mathbf{D}$, it is possible to embed the data in $\mathcal{S}$ by developing the so-called dissimilarity space representation: each input sample is represented by the corresponding row-vector in $\mathbf{D}$.

Recently, the Optimized Dissimilarity Space Embedding (ODSE) system has been proposed as a general labeled graph classifier, achieving state-of-the-art results in terms of classification accuracy on well-known benchmarking datasets \cite{odse}. The synthesis of the ODSE classification model is performed by a novel information-theoretic interpretation of the DM in terms of \textit{conveyed information}.
In practice, the system estimates the informativeness of the input data dissimilarity representation by calculating the quadratic R\'{e}nyi entropy (QRE) \cite{principe2010}.
Such an entropic characterization has been used in the compression--expansion scheme as well as an important factor of the ODSE objective function.
However, deriving the ODSE classification model is computationally demanding. As a consequence, we have developed two improved versions of the ODSE graph classification system \cite{odse2_ijcnn_2013}, which are based on a fast clustering-based compression (CBC) scheme.
The parameters of such a clustering algorithm are analytically determined, causing a considerable computational speed-up of the model synthesis phase, yet maintaining state-of-the-art standards in terms of test set classification accuracy.

In this paper, we elaborate further over the same CBC scheme first introduced in \cite{odse2_ijcnn_2013} by estimating the differential $\alpha$-order R\'{e}nyi entropy of the DVs by means of a faster technique that relies on an entropic Minimum Spanning Tree (MST).
Also in this case, we give a formal proof pertaining the setting of the clustering algorithm governing the compression.
We experimentally demonstrate that the performance of ODSE operating with the MST-based estimator is comparable with the one using the kernel-based estimator.
Additionally, we observe that with the former the overall computing time is in general lower.

The remainder of the paper is organized as follows.
In Table \ref{tab:acronyms} we report all acronyms used in this paper.
Section \ref{sec:background} provides the necessary theoretical background related to the entropy estimators used in this work.
In Section \ref{sec:odse} we give an overview of the original ODSE graph classification system design \cite{odse}.
In Section \ref{sec:odse2} we present the improved ODSE system, which is primarily discussed considering the QRE estimator.
In Section \ref{sec:bsas_ordering}, we discuss a relevant topic related to the (worst-case) efficiency of the developed CBC procedure.
Section \ref{sec:mst_compression} introduces the principal theoretical contribution of this paper.
We prove a theorem related to the CBC scheme when considering the MST-based estimator.
Experiments and comparisons with other graph classifiers on well-known benchmarking datasets are presented in Section \ref{sec:exps}.
Conclusions and future directions follow in Section \ref{sec:conclusions}.
\begin{table}[thp!]\scriptsize
\begin{center}
\caption{Acronyms sorted in alphabetic order.}
\label{tab:acronyms}
\begin{tabular}{cc}
\hline
\textbf{Acronym} & \textbf{Full name} \\
\hline
BSAS & Basic sequential algorithmic scheme \\ 
CBC & Clustering-based compression \\ 
DM & Dissimilarity matrix \\ 
DS & Dissimilarity space \\ 
DV & Dissimilarity value \\ 
IGM & Inexact graph matching \\ 
MinSOD & Minimum sum of distances \\ 
MMN & Min-max network \\ 
MS & Mode Seek \\ 
MST & Minimum spanning tree \\ 
MST-RE & Minimum spanning tree - R\'{e}nyi entropy \\ 
ODSE & Optimized dissimilarity space embedding \\ 
QRE & Quadratic R\'{e}nyi entropy \\ 
RS & Representation set \\ 
SOA & State-of-the-art \\
SVM & Support vector machines \\ 
TWEC & Triple-weight edit scheme \\
\hline
\end{tabular}
\end{center}
\end{table}

\section{Differential R\'{e}nyi entropy estimators}
\label{sec:background}

Designing pattern recognition systems by using concepts derived from information theory is nowadays well-established \cite{principe2010}.
A key issue in this context is the estimation of information-theoretic quantities from a given dataset, such as entropy and mutual information.
From the groundbreaking work of Shannon, different generalized entropy formulations have been proposed.
Here we are interested in the generalization proposed by R\'{e}nyi, which is called $\alpha$-order R\'{e}nyi entropy.
Given a continuous random variable $X$, distributed according to a probability density function $p(\cdot)$, the $\alpha$-order R\'{e}nyi entropy is defined as:
\begin{equation}
\label{eq:differential_entropy}
H_{\alpha}(X)=\frac{1}{1-\alpha}\log\left(\int p(x)^{\alpha}dx\right), \ \alpha\geq0, \alpha\neq1.
\end{equation}

In the following two subsections, we provide the details of the non-parametric $\alpha$-order R\'{e}nyi entropy estimation techniques used here.

\subsection{The QRE estimator}
\label{sec:ip_entropy}

Recently, Pr{\'i}ncipe \cite{principe2010} provided a formulation of Eq. \ref{eq:differential_entropy} in terms of the so-called \textit{information potential} of order $\alpha$, $V_{\alpha}(X)$,
\begin{align}
\label{eq:principe_renyi_entropy}
V_{\alpha}(X) = \int p(x)^{\alpha}dx; \ \ H_{\alpha}(X) = -\log\left( V_{\alpha}(X)^{\frac{1}{\alpha-1}} \right) .
\end{align}

When $\alpha=2$, Eq. \ref{eq:principe_renyi_entropy} simplifies to the so-called quadratic R\'{e}nyi entropy.
Non-parametric kernel-based estimators provide a plug-in solution for the density estimation problem.
Typically, a zero-mean Gaussian kernel $G_{\sigma}(\cdot)$ is adopted, $\widetilde{p}(x)=\frac{1}{n}\sum_{i=1}^{n} G_{\sigma}\left(x - x_i \right) $.
The Gaussian kernel $G_{\sigma}(\cdot)$ enables a controllable bias--variance trade-off of the estimator dependent on the \textit{kernel size} $\sigma$ (and on the data sample size $n$).
According to Pr{\'i}ncipe \cite{principe2010}, the QRE of the joint distribution of a \textit{d}-dimensional random vector can be estimated by relying on $d$ different unidimensional kernel estimators combined as follows:
\begin{equation}
\label{eq:joint_estimator}
\widetilde{V}_{2,\sigma}(\underline{\mathbf{X}}_n)=\frac{1}{n^{2}} \displaystyle\sum_{i=1,j=1}^{n} \left( \prod_{r=1}^{d}G_{\sigma\sqrt{2}} \left( x_{j}^{(r)}-x_i^{(r)} \right) \right) ,
\end{equation}
where $\widetilde{V}_{2,\sigma}(\cdot)$ is the quadratic information potential and $G_{\sigma\sqrt{2}}(\cdot)$ is a convoluted Gaussian kernel with doubled variance, evaluated at the difference between the realizations.
Since the input domain is bounded, the entropy is maximized when the distribution is uniform, $\max H_{2}(X) = d\times \log(\Delta)$, where $\Delta$ is the input data extent \cite{principe2010}.

$O(dn^2)$ kernel evaluations are needed to compute (\ref{eq:joint_estimator}), which may become onerous due to the cost of computing the exponential function.

\subsection{The MST-based estimator}
\label{sec:mst_compression_desc}

Let $\underline{\mathbf{X}}_{n}$ be the data sample of $n$ measurements (points), with $\underline{\mathbf{x}}_i\in\mathbb{R}^{d}, i=1, 2, ..., n$, and $d\geq 2$, and let $G(\underline{\mathbf{X}}_{n})$ be the complete (entropic) graph constructed over these $n$ measurements.
An edge $e_{ij}$ of such a graph connects $\underline{\mathbf{x}}_i$ and $\underline{\mathbf{x}}_j$ in $\mathbb{R}^d$ by means of a straight line described by the length $|e_{ij}|$, which is computed taking the Euclidean distance:
\begin{equation}
|e_{ij}| = d_{2}(\underline{\mathbf{x}}_i, \underline{\mathbf{x}}_j) .
\end{equation}

The $\alpha$-order R\'{e}nyi entropy (\ref{eq:differential_entropy}) can be estimated according to a geometric interpretation of a MST of $G(\underline{\mathbf{X}}_{n})$ in $\mathbb{R}^d$ (shortened as MST-RE).
To this end, let $L_{\gamma}(\underline{\mathbf{X}}_{n})$ be the \textit{weighted length} of a MST $T$ connecting the $n$ points, which is defined as
\begin{equation}
\label{eq:l_mst}
L_{\gamma}(\underline{\mathbf{X}}_{n}) = \displaystyle\min_{T\in\mathcal{T}(G(\underline{\mathbf{X}}_{n}))} \displaystyle\sum_{e_{ij}\in T} |e_{ij}|^{\gamma} ,
\end{equation}
where $\gamma\in(0, d)$ is a user-defined parameter, and $\mathcal{T}(G(\underline{\mathbf{X}}_{n}))$ is the set of all possible (entropic) spanning trees of $G(\underline{\mathbf{X}}_{n})$.
The R\'{e}nyi entropy of order $\alpha\in(0, 1)$, elaborated using the MST length (\ref{eq:l_mst}), is defined as follows \cite{bonev__2008,Hero_Asympt__1999}:
\begin{equation}
\label{eq:rentropy_mst}
\hat{H}_{\alpha}(\underline{\mathbf{X}}_{n}) = \frac{d}{\gamma}\left[ \ln\left(\frac{L_{\gamma}(\underline{\mathbf{X}}_{n})}{n^{\alpha}}\right) - \ln\left(\beta(L_{\gamma}, d)\right) \right],
\end{equation}
where the order $\alpha$ is determined by calculating:
\begin{equation}
\label{eq:alpha}
\alpha=\frac{d-\gamma}{d}.
\end{equation}

The $\beta(L_{\gamma}, d)$ term is a constant (given the data dimensionality) that can be approximated, for large enough dimensions, $d$, as:
\begin{equation}
\label{eq:beta_approx}
\beta(L_{\gamma}, d) \simeq \frac{\gamma}{2}\ln\left( \frac{d}{2\pi e} \right) .
\end{equation}

By modifying $\gamma$ we obtain different $\alpha$-order R\'{e}nyi entropies.
By definition of $G(\underline{\mathbf{X}}_{n})$, MST-RE (\ref{eq:rentropy_mst}) is not sensitive to the input dimensionality.

Assuming to perform the estimation on a set of $n$ measurements in $\mathbb{R}^d$, the computational complexity involved in computing Eq. \ref{eq:rentropy_mst} is given by:
\begin{align}
\label{eq:rentropy_mst_complexity}
O\left( \frac{n(n-1)}{2}e + \frac{n(n-1)}{2}\times\log\left(\frac{n(n-1)}{2}\right) + (n-1) \right).
\end{align}

The first term in (\ref{eq:rentropy_mst_complexity}) accounts for the generation of $G(\underline{\mathbf{X}}_{n})$, computing the respective Euclidean distances for the edge weights.
The second term quantifies the cost involved in the MST computation using the well-known Kruskal's algorithm. The last term in (\ref{eq:rentropy_mst_complexity}) concerns the computation of the MST length.

\section{The original ODSE graph classifier}
\label{sec:odse}

The ODSE graph classification system \cite{odse} is founded on an explicit graph embedding mechanism that represents the input set of graphs $\mathcal{S}, n=|\mathcal{S}|$, using a suitable RS $\mathcal{R}, d=|\mathcal{R}|$, by initially computing the corresponding DM, $\mathbf{D}^{n\times d}$.
The configuration of the embedding vectors representing the input data in $\mathcal{D}$ is derived directly using the rows of $\mathbf{D}$.
The adopted IGM dissimilarity measure is the symmetric version of the procedure called best matching first that uses a three-weight edit scheme (TWEC). Although TWEC provides a heuristic solution to the graph edit distance problem, it has shown a good compromise between computational complexity (quadratic in the graph order) and the number of characterizing parameters \cite{gm_survey,odse,gralg_2012}.
TWEC performs a greedy assignment of the vertices among the two input graphs on the base of the corresponding labels dissimilarity; edge operations are induced accordingly.

ODSE synthesizes the classification model optimizing the DS representation by means of two dedicated operations, called \textit{compression} and \textit{expansion}. Both operations make use of the QRE estimator (Sec. \ref{sec:ip_entropy}) to quantify the information conveyed by the DM.

Another important component of the ODSE graph classification system is the feature-based classifier, which operates directly in $\mathcal{D}$; its own classification model is trained during the ODSE synthesis.
Such a classifier can be any well-known classification system, such as an MMN \cite{rizzi2002}, or a kernelized support vector machine (SVM).
Test labeled graphs are classified by ODSE feeding the corresponding dissimilarity representation to the learned feature-based classifier, which assigns proper class labels to the test patterns.

\begin{figure*}[ht!]
\centering
\subfigure[Training of ODSE.]{
   \includegraphics[scale=0.39,keepaspectratio=true]{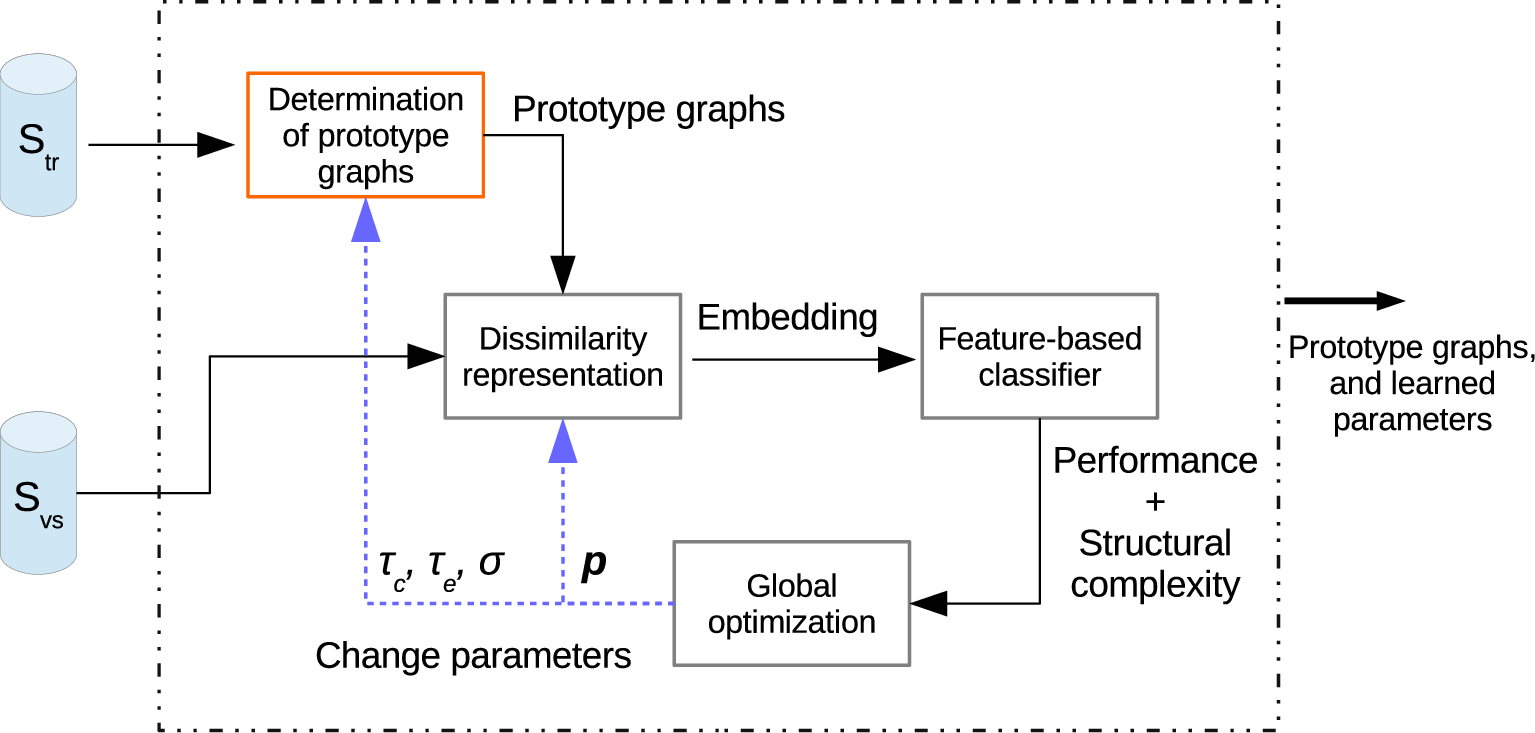}
   \label{fig:odse_training}
 }
~
 \subfigure[Determination of the prototypes.]{
   \includegraphics[scale=0.45,keepaspectratio=true]{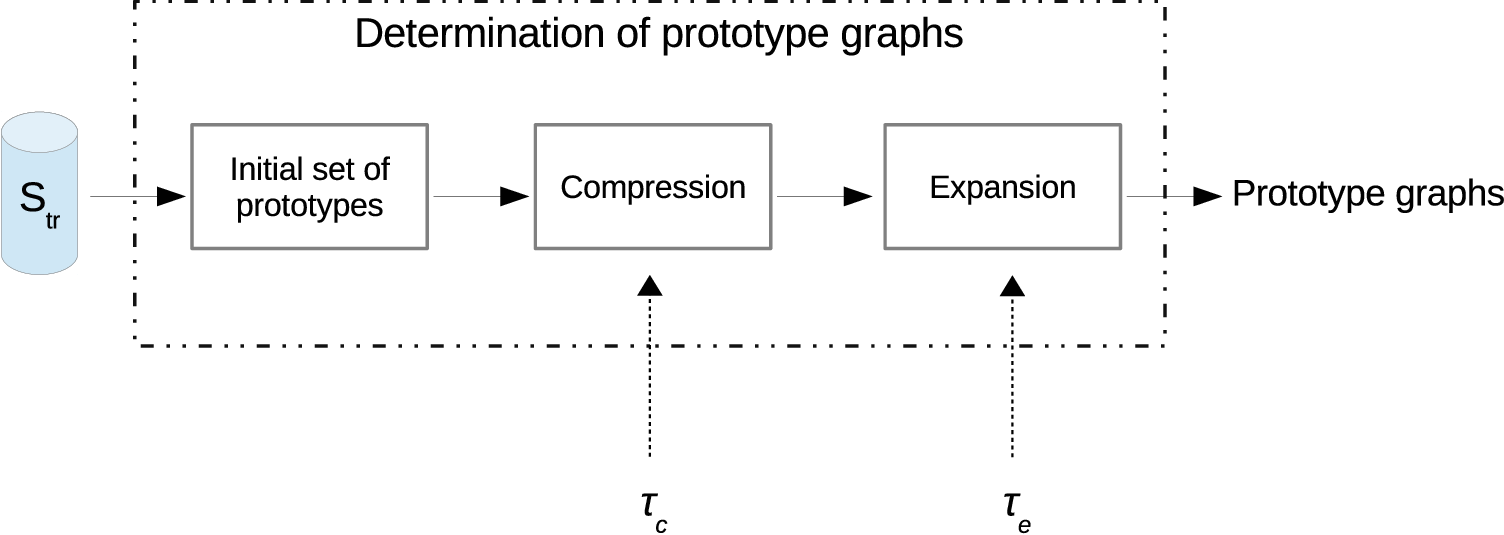}
   \label{fig:odse_prototypes}
 }
\caption{Schematic descriptions of the main stages in the ODSE training.}
\label{fig:odse}
\end{figure*}

Figs. \ref{fig:odse_training} and \ref{fig:odse_prototypes} give, respectively, the schematics of the ODSE training and determination of the prototypes.
The ODSE classification model is defined by the RS, $\mathcal{R}_i$, the TWEC parameters, $\mathbf{p}$, and the model of the trained feature-based classifier.
During the synthesis stage additional parameters are optimized: the kernel size $\sigma$ used by the entropy estimator and two thresholds, $\tau_c, \tau_e$, which are used in the compression and expansion operations, respectively.
The ODSE model is synthesized by cross-validating the learned models on the training set $\mathcal{S}_{tr}$ over a suitable validation set $\mathcal{S}_{vs}$. The global optimization is governed by a genetic algorithm, since the recognition performance guides, and its analytical definition with respect to (w.r.t.) the model parameters is not available in closed form.
The genetic algorithm, although it does not assure convergence towards a global optimum, it is easily and effectively parallelizable, allowing to make use of multicore hardware/software implementations during the training stage.

\subsection{The ODSE objective function}
\label{sec:odse_of}

All parameters characterizing the ODES model are arranged into codes, $\underline{\mathbf{c}}_i\in\mathcal{C}$. These include the two entropy thresholds $\{\tau_c, \tau_e\}_i$, the kernel size of the entropy estimator, $\{\sigma\}_i$, the weights of TWEC and any parameter of the vertex/edge label dissimilarity measures, all ranging in $[0, 1]$.
Since each $\underline{\mathbf{c}}_i$ induces a specific RS, $\mathcal{R}_i$, the optimization problem that characterizes the ODSE synthesis consists in deriving the best-performing RS:
\begin{equation}
\label{eq:opt2}
\hat{\mathcal{R}} = \displaystyle\argmax_{\underline{\mathbf{c}}_i\in\mathcal{C}} f(\mathcal{S}_{tr}, \mathcal{S}_{vs}, \mathcal{R}_i) .
\end{equation}

The objective function (\ref{eq:opt2}) is defined as a linear convex combination of two objectives,
\begin{equation}
\begin{aligned}
\label{eq:objective_func}
f(\mathcal{S}_{tr}, \mathcal{S}_{vs}, \mathcal{R}_i) = \eta f_{1}(\Phi^{\mathcal{R}_i}(\mathcal{S}_{tr}), \Phi^{\mathcal{R}_i}(\mathcal{S}_{vs})) + (1-\eta)f_{2}(\Phi^{\mathcal{R}_i}(\mathcal{S}_{tr})) ,
\end{aligned}
\end{equation}
where $\eta\in[0, 1]$ and $\Phi^{\mathcal{R}_i}(\cdot)$ shorten the dissimilarity representation of an entire dataset using the compressed-and-expanded RS instance, $\mathcal{R}_i$. The function $f_1(\cdot, \cdot)$ evaluates the recognition rate achieved on a validation set $\mathcal{S}_{vs}$, while $f_2(\cdot)$ accounts for the quality of the synthesized classification model. Specifically,
\begin{equation}
\label{eq:f2}
f_{2}(\Phi^{\mathcal{R}_i}(\mathcal{S}_{tr}))=\varsigma\Theta + (1-\varsigma)\Upsilon ,
\end{equation}
where $\varsigma\in[0, 1]$, and $\Theta$ denotes the cost related to the number $d_i$ of prototypes.
Accordingly,
\begin{equation}
\label{eq:f2_t1}
\Theta=1-\frac{d_i-\zeta}{|\mathcal{S}_{tr}|} ,
\end{equation}
where $\zeta$ is the number of classes characterizing the classification problem at hand.
The second term, namely $\Upsilon$, captures the informativeness of the DM:
\begin{equation}
\label{eq:f2_t2}
\Upsilon=\widetilde{H}_{2}(\underline{\mathbf{D}}_{n}) .
\end{equation}

We consider the entropy factor (\ref{eq:f2_t2}) in the ODSE objective function (\ref{eq:objective_func}) to increase the spread--dispersion of the DVs, which in turn is assumed to magnify the separability of the classes.

\subsection{The ODSE compression operation}
\label{sec:odse_compression}

The compression operation searches for subsets of the initial RS, $\mathcal{R}$, which convey \textit{similar information} w.r.t. $\mathcal{S}_{tr}$; the initial RS is equal to the whole $\mathcal{S}_{tr}$ in the original ODSE.
In order to describe the mechanism behind the ODSE compression operation, we need to define when a given subset $\mathcal{B}\subseteq\mathcal{R}$ of prototypes is compressible.
Let $\mathbf{D}^{n\times d}$ be the DM corresponding to $\mathcal{S}_{tr}$ and $\mathcal{R}$, with $n=|\mathcal{S}_{tr}|$ and $d=|\mathcal{R}|$.
Basically, $\mathcal{B}$ individuates a subset of $k=|\mathcal{B}|\leq d$ columns of $\mathbf{D}$.
Let $\mathbf{D}[\mathcal{B}]^{n\times k}$ be the \textit{filtered} DM, i.e., the submatrix considering the prototypes in $\mathcal{B}$ only. We say that $\mathbf{D}[\mathcal{B}]^{n\times k}$ is compressible if
\begin{equation}
\label{eq:compression_rule}
\widetilde{H}_{2}(\underline{\mathbf{D}}_k)\leq \tau_c ,
\end{equation}
where $0\leq\tau_c\leq 1$ is the compression threshold, and $\widetilde{H}(\cdot)$ estimates the QRE of the underlying joint distribution of $\mathbf{D}[\mathcal{B}]^{n\times k}$.
In practice, the values of $\mathbf{D}[\mathcal{B}]^{n\times k}$ are interpreted as $k$ measurements of a $n$-dimensional random vector; $\underline{\mathbf{D}}_k$ is the corresponding notation that we use throughout the paper to denote a sample of $k$ random measurements elaborated from the DM.
If the measurements are concentrated around a single $n$-dimensional support point, the estimated joint entropy is close to zero. This fact allows us to use Eq. \ref{eq:compression_rule} as a systematic compression rule, retaining only a single representative prototype graph of $\mathcal{B}$.

The selection of the subsets $\mathcal{B}_i, i=1, 2, ..., p$, for the compressibility evaluation is the first important algorithmic issue to be addressed.
In the original ODSE \cite{odse}, the subset selection has been performed by means of a randomized algorithm. The computational complexity of this approach is $O\left(d^3n\right)$, which does not scale adequately as the input size grows.

\subsection{The ODSE expansion operation}
\label{sec:odse_expansion}

The expansion focuses on each single $R_j\in\overleftarrow{\mathcal{R}}$, by analyzing the corresponding columns of the compressed DM, $\mathbf{D}^{n\times d}$.
By denoting with $\mathbf{\underline{D}}_n$ the sample containing the $n$ DVs corresponding to the \textit{j}-th column of $\mathbf{D}$, we say that $R_j$ is expandable if
\begin{equation}
\label{eq:exp_rule}
\widetilde{H}_{2}(\mathbf{\underline{D}}_n)\leq\tau_e  ,
\end{equation}
where $0\leq\tau_e\leq 1$ is the expansion threshold.
Practically, the information provided by the prototype is low if the $n$ unidimensional measurements are concentrated around a single real-valued number. In such a case, the estimated entropy would be low, approaching zero as the underlying distribution becomes degenerate. Examples of such prototypes are outliers and prototype graphs that are equal in the same measure to all other graphs.
Once an expandable $R_j$ is individuated through (\ref{eq:exp_rule}), $R_j$ is substituted by extracting $\zeta$ new graphs elaborated from $\mathcal{S}_{tr}$.
Notably, those new graphs are derived by searching for recurrent subgraphs in a suitable subset of the training graphs.

Although the idea of trying to extract new features by searching for (recurrent) subgraphs is interesting, it is also very expensive in terms of computational complexity.

\section{The improved ODSE graph classifier}
\label{sec:odse2}

The improved ODSE system \cite{odse2_ijcnn_2013} is designed with the primary goal of a significant computational speed-up.
The first variant, which is presented in Sec. \ref{sec:odse2v1}, considers a simple yet fast RS initialization strategy and a more advanced compression mechanism.
The compression is grounded on a formal result discussed in Sec. \ref{sec:compression}.
The second variant of the ODSE classifier is presented in Sec. \ref{sec:odse2v2}. This version includes a more elaborated initialization of the RS, while it is characterized by the same CBC operation. The expansion operation, in both cases, has been greatly simplified.
Finally, in Sec. \ref{sec:bsas_ordering} we discuss an important fact related to the efficiency of the implemented CBC.

\subsection{ODSE with clustering-based compression operation}
\label{sec:odse2v1}

\subsubsection{Randomized representation set initialization}
\label{sec:rs_init}

The initial RS $\mathcal{R}$, that is, the RS used during the synthesis, is defined by sampling the $\mathcal{S}_{tr}$ according to a selection probability, $p$. The size of the initial RS is thus characterized by a binomial distribution, ing in average $|\mathcal{S}_{tr}|p$ graphs, with variance $|\mathcal{S}_{tr}|p(1-p)$.
Although such a selection criteria is linear in the training set size, it operates \textit{blindly} and may cause an unbalanced selection of the prototypes considering the prior class distributions. However, such a simple sampling scheme is mostly used when the available hardware cannot process the entire dataset at hand.

\subsubsection{Compression by a clustering-based subset selection}
\label{sec:compression}

The entropy measured by the QRE estimator (\ref{eq:joint_estimator}) is used to determine the compressibility of a subset of prototypes, $\mathcal{B}$. Since the entropy estimation is directly related to the DVs between the graphs of $\mathcal{B}$, we design a subset selection strategy that aggregates the initial prototypes according to their distance in the DS.
Such subsets are assured to be compressible by definition, avoiding thus the computational burden involved in the entropy estimation.

We make use of the well-known Basic Sequential Algorithmic Scheme (BSAS) clustering algorithm (see the pseudo-code of Algorithm \ref{alg:bsas}) with the aim of grouping the $n$-dimensional dissimilarity column-vectors $\underline{\mathbf{x}}_j, j=1, 2, ..., d$, with (hyper)spheres, using the Euclidean metric $d_2(\cdot, \cdot)$.
The main reason behind the use of such a simple cluster generation rule is that it is much faster than other more sophisticated approaches \cite{FilipponePR08}, and it gives full control on the generated cluster geometry through a single real-valued parameter, $\theta$.
Since $\theta$ constrains each cluster $\mathcal{B}_l$ to have a maximum intra-cluster DV (i.e., a diameter) lower or equal to $2\theta$, we can deduce analytically the value of $\theta$ considering the particular instance of the kernel size $\sigma_c$ and the entropy threshold $\tau_c$ used in Eq. \ref{eq:compression_rule}.
Accordingly, the following theorem (see \cite{odse2_ijcnn_2013} for the proof) allows us to determine a partition $P(\theta; \tau_c, \sigma_c)$ that contains clusters that are compressible by construction.
\begin{thmTheta}
\label{th:thmTheta}
The compressible partition $P(\theta; \tau_c, \sigma_c)$ obtained on a training set $\mathcal{S}_{tr}$ of $n$ graphs, is derived setting:
\begin{equation}
\label{eq:f_tau_sigma}
\theta \leq \sqrt{ \frac{\tau_c n\sigma_{c}^2\ln(2)}{2} }.
\end{equation}
\end{thmTheta}
\begin{algorithm}[h!]\footnotesize
\caption{BSAS clustering algorithm.}
\label{alg:bsas}
\begin{algorithmic}[1]
\REQUIRE $n$ input data, a dissimilarity measure $d(\cdot, \cdot)$, cluster radius $\theta$, and maximum number of clusters $Q$
\ENSURE Partition $P(\theta)$
\FOR{$i=1, 2, ..., n$}
\IF{$P(\theta)=\emptyset$}
\STATE Create a new cluster in $P(\theta)$ and define $x_i$ as the set representative
\ELSE
\STATE Get the distance value $D$ from the closest representative modeling a cluster of the current partition $P(\theta)$
\STATE $D=\displaystyle\min_{\mu_j\in P(\theta)} d(x_i, \mu_j)$
\IF{$D>\theta$ AND $|P(\theta)|< Q$}
\STATE Add a new cluster in $P(\theta)$ and define $x_i$ as the representative
\ELSE
\STATE Add $x_i$ in the \textit{j}-th cluster and update the representative element
\ENDIF
\ENDIF
\ENDFOR
\end{algorithmic}
\end{algorithm}

The optimization of parameters $\tau_c$ and $\sigma_c$, together with the proof of Theorem \ref{th:thmTheta}, allows us to search for the best level of training set compression for the problem at hand.
Algorithm \ref{alg:compression_odse2} shows the pseudo-code of the herein described compression operation.
Since the ultimate aim of the compression is to aggregate prototypes that convey similar information w.r.t. $\mathcal{S}_{tr}$, we represent a cluster using the minimum sum of distances (MinSOD) technique \cite{delvescovo+livi+rizzi+frattalemascioli2011}.
In fact, the MinSOD allows to select a single representative element $x_k\in\mathcal{B}_k$ according to the following expression:
\begin{equation}
\label{eq:minsod}
x_k=\argmin_{x_j\in \mathcal{B}_k} \sum_{x_i\in \mathcal{B}_k} d_2(x_j, x_i).
\end{equation}

Eventually, the $p$ prototype graphs, $B_i, i=1, 2, ..., p$, corresponding to the $p$ computed MinSOD elements in the DS, populate the compressed RS, $\overleftarrow{\mathcal{R}}=\{B_1, B_2, ..., B_p\}$.
\begin{algorithm}[h!]\footnotesize
\caption{Clustering-based compression algorithm.}
\label{alg:compression_odse2}
\begin{algorithmic}[1]
\REQUIRE The initial set of prototype graphs $\mathcal{R}, |\mathcal{R}|=d$, the DM $\mathbf{D}^{n\times d}$, the compression threshold $\tau_c$, and the kernel size $\sigma_c$
\ENSURE The compressed set of prototype graphs $\overleftarrow{\mathcal{R}}$
\STATE Configure BSAS setting $Q=|\mathcal{R}|$ and $\theta$ according to Eq. \ref{eq:f_tau_sigma}
\STATE Let $\mathcal{X}=(\underline{\mathbf{x}}_1, \underline{\mathbf{x}}_2, ..., \underline{\mathbf{x}}_d)$ be the (ordered) set of dissimilarity vectors elaborated from the columns of $\mathbf{D}$
\STATE Execute the BSAS on $\mathcal{X}$. Let $P(\theta; \tau_c, \sigma_c)=\{\mathcal{B}_1, \mathcal{B}_2, ..., \mathcal{B}_p\}$ be the obtained compressible partition
\STATE Compute the MinSOD element $\underline{\mathbf{b}}_i$ of each cluster $\mathcal{B}_i, i=1, 2, ..., p$, according to Eq. \ref{eq:minsod}. Retrieve from $\mathcal{R}$ the prototype graph $B_i$ corresponding to each dissimilarity vector $\underline{\mathbf{b}}_i$
\STATE Define $\overleftarrow{\mathcal{R}}=\bigcup_{i=1}^{p} B_i$
\RETURN $\overleftarrow{\mathcal{R}}$
\end{algorithmic}
\end{algorithm}

The search interval for the kernel size $\sigma_c$ can be effectively reduced as follows:
\begin{equation}
\label{eq:max_sigma}
0 \leq \sigma_c \leq \sqrt{\frac{8}{\ln(2)}}.
\end{equation}

A proof for (\ref{eq:max_sigma}) can be found in \cite{odse2_ijcnn_2013}. This bound is important, since it allows to narrow the search interval for the kernel size $\sigma_c$, which is theoretically defined in the entire extended real line.

\subsubsection{Expansion based on replacement with maximum dissimilar graphs}
\label{sec:odse2_expansion}

The genetic algorithm evolves a population of models over the iterations $t=1, 2, ..., \mathrm{max}$. Let $\mathcal{R}^{0}$ be defined as shown in Sec. \ref{sec:rs_init}, and let $\mathcal{N}^{t}=\mathcal{S}_{tr}\setminus\mathcal{R}^{t-1}$ be the set of unselected training graphs at iteration $t\geq 1$.
Finally, let $\overleftarrow{\mathcal{R}}^{t}$ be the compressed RS at iteration $t$. The herein described expansion operation makes use of the elements of $\mathcal{N}^{t}$ replacing in $\overleftarrow{\mathcal{R}}^{t}$ those prototypes that do not discriminate the classes.
The check for the expansion of a single prototype graph is still performed as described in Sec. \ref{sec:odse_expansion}.
Notably, if the estimated entropy from the $j$-th column vector is lower than the expansion threshold, $\tau_e$, then $l$ new training graphs are selected from $\mathcal{N}^t$ for each class, where $l\geq 1$ is user-defined.
Those $\zeta\times l$ new graphs are selected such that they result maximally dissimilar w.r.t. the $j$-th prototype under analysis.
The new expansion procedure is outlined in \cite[Algorithm~2]{odse2_ijcnn_2013}.

Since compression and expansion are evaluated considering two different interpretations of the DM, we accordingly use two different kernel sizes: $\sigma_c$ and $\sigma_e$.

\subsubsection{Analysis of computational complexity}
\label{sec:comp_complexity_odse2}

The computational complexity is dictated by the execution of the genetic algorithm, $O(I + EP\times F)$.
$I$ is the cost of the RS initialization, $E$ is the number of (maximum) evolutions, $P$ is the population size, and finally $F$ is the cost related to a single fitness function evaluation.
In this system variant, the initialization is linear in the training set size, $O(I) = O(|\mathcal{S}_{tr}|)$; in average we select $d^{'}= \lfloor|\mathcal{S}_{tr}|p \rfloor$ prototypes.
The detailed cost related to the fitness function, $O(F)$, is articulated as the sum of the following costs:
\begin{equation}
\label{eq:fitness_cost_odse2}
\begin{aligned}
O(F_1)&=O(nd^{'}g); \ O(F_2)= O(nQCe); \\
O(F_3)&=O\left( \overleftarrow{d}n^2\times\left( N\log(N) + \zeta l \right) \right); \\
O(F_4) &= O\left(n\overline{d}\right); \ \ O(F_5) = O\left(v\times\left(\overline{d}+kn\right)\right); \\
O(F_6) &= O\left(n^2\overline{d}\right).
\end{aligned}
\end{equation}

The first cost, $F_1$, is related to the computation of the initial DM corresponding to $\mathcal{S}_{tr}$ with RS obtained through the initialization of Sec. \ref{sec:rs_init}; $g$ is the computational cost associated with the adopted IGM procedure.
$F_2$ is due to the compression operation which consists in a single BSAS execution, where $C=d^{'}$ is the cache size of the MinSOD \cite{delvescovo+livi+rizzi+frattalemascioli2011}, $Q=d^{'}$, and $e=n$ is the cost of a single Euclidean distance computation.
$F_3$ is the cost characterizing the expansion operation; $N$ is the cardinality of the set $\mathcal{N}^{t}$. This operation is repeated at most $\overleftarrow{d}=|\overleftarrow{\mathcal{R}}|$ times, with a quadratic entropy estimation cost in the training set size.
$F_4$ is the cost related to the embedding of the DM, and $F_5$ is due to the classification of the validation set using a \textit{k}-NN rule based classifier -- this cost is updated according to the specific classifier.
$F_6$ is the cost for the QRE over the compressed-and-expanded DM.

As it is possible to deduce from Eq. \ref{eq:fitness_cost_odse2}, the model synthesis is now characterized by a quadratic cost in the training set size, $n$, as well as in the RS size, $d$, while in the original ODSE it was (pseudo) cubic in both $n$ and $d$.

\subsection{ODSE with mode seeking initialization}
\label{sec:odse2v2}

The ODSE version described here does not include any expansion operation. The RS initialization is now part of the synthesis, since it depends on some of the parameters tuned during the optimization. Compression is still implemented as described in Sec. \ref{sec:compression}.

The initialization makes use of the Mode Seek (MS) algorithm \cite{pkekalska+duin2005}, which is a well-known procedure that is able to individuate the \textit{modes} of a distribution. For each class $c_i, i=1, 2, ..., \zeta$, and considering a user-defined neighborhood size $s\geq1$, the algorithm proceeds as illustrated in \cite[Algorithm~3]{odse2_ijcnn_2013}. The elements of $\mathcal{R}$ found in this way are the estimated modes of the class distribution; hence it is a supervised algorithm.
The cardinality of $\mathcal{R}$ depends on the choice of $s$: the larger is $s$, the smaller $\mathcal{R}$. This approach is very appropriate when elements of the same class are distributed in different and heterogeneous clusters: the cluster representatives are the modes individuated by the MS algorithm.
Moreover, the MS algorithm can be useful to filter out outliers, since they are characterized by a low neighborhood density.
The procedure depends on $s$, which directly influences the outcome of the initialization. Additionally, since the neighborhood is defined in the graph domain, MS is also dependent on the weights characterizing TWEC (in our case). For this very reason, the initialization is now performed during the ODSE synthesis.

To limit the complexity of such an initialization, in the experiments we systematically assign small values to $s$, constraining the search in small neighborhoods. A possible side effect of this choice is that we can find an excessive number of prototypes/modes. This effect is however attenuated by the compression algorithm (\ref{alg:compression_odse2}).

\subsubsection{Analysis of computational complexity}
\label{sec:comp_complexity_odse2.2}

The overall computational cost of the synthesis is now bounded by $O(EP\times F)$; see (\ref{eq:fitness_cost_odse2.2}).
The two main steps of the fitness function involve the execution of the MS algorithm followed by the compression algorithm.
The $F_1$ cost refers to the MS algorithm. $|c_i|$ is the number of training data belonging to the \textit{i}-th class.
$F_2$ refers to the computation of the initial DM, constructed using $\mathcal{S}_{tr}$ and the $d^{'}\leq|\mathcal{S}_{tr}|$ prototypes derived with MS.
$F_3$ is the cost of the compression operation, with $Q=d^{'}$. $F_4$, $F_5$, and $F_6$ are equivalent to the ones described in Sec. \ref{sec:comp_complexity_odse2.2}.
The overall cost is dominated by the initialization stage (the $F_1$ cost), which is (pseudo) quadratic in the class size $|c_i|$, and quadratic in the neighborhood size, $s$.
\begin{align}
\label{eq:fitness_cost_odse2.2}
\nonumber O(F_1)&=O\left( n+\zeta |c_i|\times\left(|c_i|g+|c_i|\log(|c_i|)+s+s^2 \right) \right); \\
O(F_2)&=O(nd^{'}g); \ \ O(F_3)= O(nQCe); \\
\nonumber O(F_4)&=O\left(n\overline{d}\right); \ \ O(F_5)=O\left(v\times\left(\overline{d}+kn\right)\right); \\
\nonumber O(F_6)&=O\left(n^2\overline{d}\right).
\end{align}

\subsection{The efficiency of the ODSE clustering-based compression}
\label{sec:bsas_ordering}

BSAS (see Algorithm \ref{alg:bsas}) is characterized by a linear computational complexity.
However, due to the sequential processing nature, the outcome is sensitive to the data presentation order.
In the following, we study the effect caused by the ordering of the input over the effectiveness of the CBC, by calculating what we called ODSE \textit{compression efficiency} factor.

Let $s=(\underline{\mathbf{x}}_1, \underline{\mathbf{x}}_2, ..., \underline{\mathbf{x}}_n)$ be the sequence of dissimilarity vectors describing the $n$ prototypes in the DS, which are presented in input to Algorithm \ref{alg:bsas}. Let $\Omega(s)$ be the set of all permutations of the sequence $s$. We define the optimal compression ratio $\rho^{*}(s)$ for the sequence $s$ as:
\begin{equation}
\rho^{*}(s) = \max_{s_i\in\Omega(s)} \rho(s_i) = \max_{s_i\in\Omega(s)} |\mathcal{R}|/|\overleftarrow{\mathcal{R}}_i| ,
\end{equation}
where $\overleftarrow{\mathcal{R}}_i$ is the compressed RS obtained by analyzing the prototypes arranged according to $s_i$, and $\mathcal{R}$ is the uncompressed RS, i.e., the initial RS.
Let $\hat{\rho}(s)$ be the effective compression ratio, achieved by ODSE considering a generic ordering of $s$. The ratio
\begin{equation}
\label{eq:efficiency}
\xi = \lim_{n\rightarrow\infty} \hat{\rho}(s) / \rho^{*}(s) \in [0, 1],
\end{equation}
describes the asymptotic efficiency of the ODSE compression as the initial RS size grows.
\begin{thmSequenceOrdering}
\label{th:thmSequenceOrdering}
The asymptotic worst-case ODSE compression efficiency factor is $\xi=2/3$.
\end{thmSequenceOrdering}

The proof can be found in Appendix \ref{sec:appendixA}.
An interpretation of the result of Theorem \ref{th:thmSequenceOrdering} is that, in the general case, the asymptotic efficiency of the implemented CBC varies within the $[2/3, 1]$ range of the optimum compression.

\section{ODSE with the MST-based R\'{e}nyi entropy estimator}
\label{sec:mst_compression}

In the following, we contextualize the MST-RE estimation technique introduced in Sec. \ref{sec:mst_compression_desc} as a component of the improved ODSE system presented in Sec. \ref{sec:odse2}.
Notably, we provide a theorem for determining the $\theta$ parameter of BSAS used in the compression operation (Algorithm \ref{alg:compression_odse2}).
In this case, we generate clusters according to the particular instance of $\tau_c$ and of the $\gamma$ parameter, since the kernel size parameter, $\sigma_c$, is not present in the MST-based estimator.
The $\gamma$ parameter is optimized during the ODSE synthesis. While $\gamma$ is defined in $(0, d)$, where $d$ is the dimensionality of the samples, we restrict the search interval to $(0, U]$, with $U=3$ in the experiments.
This technical choice is motivated by the fact that $\gamma$ is used in Eq. \ref{eq:l_mst} as exponent, and an excessively large value would easily cause \textit{overflow} problems of the MST length variable floating-point representation.
\begin{thmThetaMST}
\label{th:thmThetaMST}
Considering the instances of $\gamma$ and $\tau_c$, the compressible partition $P(\theta; \tau_c, \gamma)$ is derived executing the BSAS algorithm on $n=|\mathcal{S}_{tr}|$ training graphs by setting:
\begin{equation}
\label{eq:theta_bound_mst}
\theta \leq 2^{\tau_c-1} n^{\frac{\tau_c}{2}} \beta^{\frac{-\tau_c+1}{\gamma}} c(\gamma),\ \mathrm{where }\ 0 \leq c(\gamma) \leq 2^{\frac{\alpha}{\gamma}} .
\end{equation}
\end{thmThetaMST}

The proof of this theorem can be found in Appendix \ref{sec:appendixB}.
Defining $\theta$ according to Eq. \ref{eq:theta_bound_mst} constrains the BSAS to generate clusters that are compressible by construction.
Since $\tau_c$ and $\gamma$ are optimized during the synthesis of the classifier, the result of Theorem \ref{th:thmThetaMST}, likewise the one of Theorem \ref{th:thmTheta}, allows us to evaluate different levels of training set compression according to the overall system performance.
It goes without saying that computational complexity discussed in the previous sections is readily updated by considering the cost of the MST-based estimator (see Eq. \ref{eq:rentropy_mst_complexity}).

\section{Experiments}
\label{sec:exps}

In Sec. \ref{sec:datasets} we introduce the IAM benchmarking datasets.
In Sec. \ref{sec:setting} we present experimental setting.
Finally, in Sec. \ref{sec:results} we show and discuss the results.

\subsection{Datasets}
\label{sec:datasets}

The experimental evaluation is performed on the well-known IAM graph benchmarking databases \cite{riesen+bunke2008}.
The IAM repository contains many different datasets representing real-world data collected from various fields: from images to biochemical compounds. In particular, we use the \emph{Letter LOW} (L-L), \emph{Letter MED} (L-M), \emph{Letter HIGH} (L-H), \emph{AIDS} (AIDS), \emph{Proteins} (P), \emph{GREC} (G), \emph{Mutagenicity} (M), and finally the \textit{Coil-Del} (C-D) datasets.
The first three are datasets of digitized characters modeled as labeled graphs, which are characterized by three different levels of noise.
The AIDS, P, and M datasets represent biochemical networks, while G and C-D are images of various type.
For the sake of brevity, we report only essential details in Tab. \ref{tab:iam_details}, referring the reader to Ref. \cite{riesen+bunke2008} (and references therein) for a more in-depth discussion about the data. Moreover, since each dataset contains graphs characterized by different vertex and edge labels, we adopted the same vertex and edge dissimilarity measures described in \cite{odse,gralg_2012}.
\begin{table}[tph!]\scriptsize
\caption{IAM datasets. See \cite{riesen+bunke2008} for details.}
\begin{center}
\begin{tabular}{ccccc}
\hline
\textbf{DS} & \textbf{\# (tr, vs, ts)} & \textbf{Classes} & \textbf{Avg. $|\mathcal{V}|$} & \textbf{Avg. $|\mathcal{E}|$} \\
\hline
L-L & (750, 750, 750) & 15 & 4.7 & 3.1 \\
L-M & (750, 750, 750) & 15 & 4.7 & 3.2 \\
L-H & (750, 750, 750) & 15 & 4.7 & 4.5 \\
AIDS & (250, 250, 1500) & 2 & 15.7 & 16.2 \\
P & (200, 200, 200) & 6 & 32.6 & 62.1 \\
G & (286, 286, 528) & 22 & 11.5 & 12.2 \\
M & (1500, 500, 2337) & 2 & 30.3 & 30.8 \\
C-D & (2400, 500, 1000) & 100 & 21.5 & 54.2 \\
\hline
\end{tabular}
\end{center}
\label{tab:iam_details}
\end{table}

\subsection{Experimental setting}
\label{sec:setting}

The ODSE system version described in Sec. \ref{sec:odse2v1} is denoted as ODSE2v1, while the version described in Sec. \ref{sec:odse2v2} as ODSE2v2.
These two versions make use of the QRE estimator; the setting of the clustering algorithm parameter $\theta$ used during the compression is hence performed according to the result of Theorem \ref{th:thmTheta}.
By following the same algorithmic scheme, we consider two additional ODSE variants that differ only in the use of the MST-RE estimator. We denote those two variants as ODSE2v1-MST and ODSE2v2-MST.
The setting of $\theta$ is hence performed according to the proof of Theorem \ref{th:thmThetaMST}.
However, the MST-based estimator is conceived for high-dimensional data. As a consequence, in the ODSE2v1-MST system version we still use the QRE estimator in the expansion operation.
We adopted two core classifiers operating in the DS. The first one is a \textit{k}-nearest neighbors (\textit{k}-NN) rule based classifier equipped with the Euclidean distance, testing three values of $k$: 1, 3, and 5. We also consider a fast MMN, which is trained with the ARC algorithm \cite{rizzi2002}.
The four aforementioned ODSE variants (i.e., ODSE2v1, ODSEv2, ODSEv1-MST, and ODSE2v2-MST) are therefore replicated into additional four variants that are straightforwardly denoted as ODSE2v1-MMN, ODSEv2-MMN, ODSEv1-MST-MMN, and ODSE2v2-MST-MMN, meaning that we just use the neuro-fuzzy MMN on the embedding space, instead of the \textit{k}-NN.
Tab. \ref{tab:odse2_conf} summarizes all ODSE configurations evaluated in this paper.

Tests are executed setting the genetic algorithm with a (fixed) population size of 30 individuals, and performing a maximum of 40 evolutions for the synthesis; a check on the fitness value is however performed terminating the optimization if the fitness does not change for 15 evolutions.
This setup has been chosen to allow a fair comparison with the previously obtained results \cite{odse,odse2_ijcnn_2013}.
The genetic algorithm performs roulette wheel selection, two-point crossover, and random mutation on the aforementioned codes $\underline{\mathbf{c}}_i$, encoding the real-valued model parameters; in addition, the genetic algorithm implements an elitism strategy which automatically imports the fittest individual into the next population.
In all configurations, we executed the system setting $\eta=0.9$ and $\varsigma=0.2$ in Eq. \ref{eq:objective_func} and \ref{eq:f2}, respectively. Moreover, the $s$ parameter affecting the MS algorithm has been set as follows: 10 for the L-L, L-M, and L-H, 20 for AIDS, 2 for P, 8 for G, and finally 100 for either M and C-D. Note that these values has been defined according to the training dataset sizes and considering some preliminary tests.
Each dataset has been processed five times using different random seeds, reporting hence the average test set classification accuracy together with its standard deviation. We report also the required average serial CPU time and the average RS size obtained after the synthesis. Tests have been conducted on a regular desktop machine with an Intel Core2 Quad CPU Q6600 at 2.40GHz and 4Gb of RAM; software is implemented in C++ on a Linux operating system using the SPARE library \cite{spare_graph_2013}. Finally, the computing time is measured using the \textit{clock()} routine of the standard \textit{ctime} library.
\begin{table*}[tph!]\scriptsize
\caption{Summary of the ODSE configurations evaluated in the experiments. The ``Init'' column refers to the RS initialization scheme, ``Compression / Est.'' refers to the compression algorithm and adopted entropy estimator, ``Expansion / Est.'' the same but for the expansion algorithm, and ``Obj. Func. (\ref{eq:f2_t2})'' refers to the entropy estimator adopted in Eq. \ref{eq:f2_t2}. Finally, ``FB Class.'' specifies the feature-based classifier operating in the DS.}
\begin{center}
\begin{tabular}{cccccc}
\hline
\textbf{Acronym} & \textbf{Init} & \textbf{Compression / Est.} & \textbf{Expansion / Est.} & \textbf{Obj. Func.} (\ref{eq:f2_t2}) & \textbf{FB Class.} \\
\hline
ODSE2v1 & Sec. \ref{sec:rs_init} & Sec. \ref{sec:compression} / QRE & Sec. \ref{sec:odse2_expansion} / QRE & QRE & \textit{k}-NN \\
ODSE2v2 & Sec. \ref{sec:odse2v2} & Sec. \ref{sec:compression} / QRE & -- & QRE & \textit{k}-NN \\
ODSE2v1-MST & Sec. \ref{sec:rs_init} & Sec. \ref{sec:compression} / MST-RE & Sec. \ref{sec:odse2_expansion} / QRE & MST-RE & \textit{k}-NN \\
ODSE2v2-MST & Sec. \ref{sec:odse2v2} & Sec. \ref{sec:compression} / MST-RE & -- & MST-RE & \textit{k}-NN \\
ODSE2v1-MMN & Sec. \ref{sec:rs_init} & Sec. \ref{sec:compression} / QRE & Sec. \ref{sec:odse2_expansion} / QRE & QRE & MMN \\
ODSE2v2-MMN & Sec. \ref{sec:odse2v2} & Sec. \ref{sec:compression} / QRE & -- & QRE & MMN \\
ODSE2v1-MST-MMN & Sec. \ref{sec:rs_init} & Sec. \ref{sec:compression} / MST-RE & Sec. \ref{sec:odse2_expansion} / QRE & MST-RE & MMN \\
ODSE2v2-MST-MMN & Sec. \ref{sec:odse2v2} & Sec. \ref{sec:compression} / MST-RE & -- & MST-RE & MMN \\
\hline
\end{tabular}
\end{center}
\label{tab:odse2_conf}
\end{table*}

\subsection{Results and discussion}
\label{sec:results}

All test set classification accuracy results have been collected in Tab. \ref{tab:results_emb_iam}. These include the results of three baseline reference systems and several state-of-the-art (SOA) classification systems based on graph embedding techniques.
The table is divided in appropriate macro blocks to simplify the comparison of the results.
The three reference systems are denoted as RPS+TWEC+\textit{k}-NN, \textit{k}-NN+TWEC, and RPS+TWEC+MMN. The first one performs a (class-independent) randomized selection of the training graphs to develop the dissimilarity representation of the input data. This system adopts the same TWEC used in ODSE and performs the classification in the DS by means of a \textit{k}-NN classifier equipped with the Euclidean distance.
The second one differs from the first system by using instead the MMN.
Finally, the third reference system operates directly in $\mathcal{G}$ by means of a \textit{k}-NN rule based classifier equipped with TWEC.
In all cases, to obtain a fair comparison with ODSE, the configuration of the dissimilarity measures for the vertex/edge labels is consistent with the one adopted for ODSE. Additionally, $k=1, 3$, and $5$ is used in the \textit{k}-NN rule, performing the TWEC parameters optimization (i.e., the weighting parameters in $[0, 1$]) by means of the same aforementioned genetic algorithm implementation. 
Therefore, also in this case the test set results must be intended as the average of five different runs (however we omit standard deviations for the sake of brevity).

Tab. \ref{tab:results_emb_iam} presents the obtained test set classification accuracy results, while Tab. \ref{tab:results_emb_iam_sd} gives the corresponding standard deviations.
We provide two types of statistical evaluation of such results. First, we perform pairwise comparisons by means of \emph{t}-test; we adopt the usual 5\% as significance threshold. Notably, we check if any of the improved ODSE variants significantly outperforms, for each dataset, both the reference systems and original ODSE. Best results satisfying such a condition are reported in bold in Tab. \ref{tab:results_emb_iam}. In addition to the pairwise comparisons, we calculate also a global ranking of all classifiers by means of the Friedman test. Missing values are replaced by the dataset-specific averages.

First of all, we note that results obtained with the baseline reference systems are always worse than those obtained with ODSE.
Test set classification accuracy percentages obtained by ODSE2v1-MST and ODSE2v2-MST are comparable with those of ODSE2v1 and ODSE2v2, although we note a slightly general improvement for the first two variants.
Results are also more stable varying the neighborhood size parameter, $k$, of the \textit{k}-NN rule. It is worth noting that, for difficult datasets as P and C-D, increasing the neighborhood size in the \textit{k}-NN rule affects significantly the test set performance (i.e., results degrade considerably).
Test set classification accuracy results obtained by means of the MMN operating in the DS are in general (slightly) inferior w.r.t. the ones obtained with the \textit{k}-NN rule -- setting $k=1$.
This result is not too unusual since the \textit{k}-NN rule is a valuable classifier, especially in absence of noisy data.
Since ODSE operates by searching for the best-performing DS for the data at hand, we may deduce that the embedding vectors are sufficiently well-organized w.r.t. the classes.
Test set results on the first four datasets (i.e., L-L, L-M, L-H, and AIDS) denote an important improvement over a large part of the SOA systems. On the other hand, results over the P, G, and M datasets are comparable w.r.t. those of the SOA systems.
For all ODSE configurations, we observe non convincing results on the C-D dataset; in this case results are comparable only with those of the reference systems (first block of Tab. \ref{tab:results_emb_iam}). However, a rational reason explaining this fact is not emerged from the tests yet, requiring thus more future investigations.
The global picture provided by the column denoted as ``Rank'' shows that the ODSE classifiers rank in general very well w.r.t. the SOA systems.
Standard deviations (Tab. \ref{tab:results_emb_iam_sd}) are reasonably small, denoting a reliable classifier regardless the particular ODSE variant.

We demonstrated that the asymptotic computational complexity of ODSE2 is quadratic, while the original ODSE was characterized by a cubic computational complexity.
Here, in order to complement this result with experimental evidence, we discuss also the effective computing time.
The calculated serial CPU time, for each dataset, is shown in Tab. \ref{tab:results_emb_iam_time}, which includes both ODSE synthesis and test set evaluation.
The ODSE variants based on the MST entropy estimator are faster, with the only exception for the P and C-D datasets. This fact is magnified on the first four datasets, in which the speed-up factor w.r.t. the original ODSE increases considerably.
The speed-up factors obtained for the first three datasets are one order of magnitude higher than the ones obtained in the other datasets.
In order to provide an explanation for such differences, we need to take a closer look at the dataset details shown in Tab. \ref{tab:iam_details}, computational complexity in Eqs. \ref{eq:fitness_cost_odse2} and \ref{eq:fitness_cost_odse2.2}, and the computational complexity of the original ODSE \cite{odse}.
It is possible to notice that the first three datasets contain smaller (in average) labeled graphs.
Therefore, this points us to look for the related terms in the computational complexity formulae.
The $g$ term (the cost of the graph matching algorithm) is directly affected by the size of the graphs and appears in $F_1$ Eq. \ref{eq:fitness_cost_odse2} and $F_1, F_2$ in Eq. \ref{eq:fitness_cost_odse2.2}. The same $g$ term appears also in $F_1$ of Eq. 24 in \cite{odse}.
In the original ODSE version \cite{odse}, the dissimilarity matrix is constructed using an initial set of prototypes equal to the training set (then it is compressed and expanded).
In the new version presented here, we instead use a reduced set with $d^{'}$ elements. In the first variant that we presented, $d^{'}$ graphs are selected randomly from the training set based on a selection probability. In the second variant, instead, we use the MS algorithm, which finds a much lower number of representatives (although, as said in the experimental setting section, we use a conservative setting for MS).
This fact provides a first rational justification for explaining the aforementioned differences. In fact, graph matching algorithms are expensive from the computational viewpoint (the adopted algorithm is quadratic in the number of vertices).
In addition, compression and expansion operations are now much faster (from cubic to quadratic in time). As shown in Tab. \ref{tab:results_emb_iam_rs}, the new ODSE versions compute a smaller RS; a direct consequence of the improved compression operation.
This is another important factor contributing to the overall speed-up, since smaller RSs imply less graph matching computations during the validation and test stages (we remind that ODSE is trained by cross-validation).
Clearly, there are also other factors, such as the convergence of the optimization algorithm, which might be affected by the specific dataset at hand.

As expected, the speed-up factors obtained by using the MMN as classifier are in general higher than those obtained with kNN.
In fact, the MMN synthesizes a classification model over the training data embedded into a DS. This significantly reduces the computing time necessary for the evaluation of the test set (and also of the validation stage performed during the synthesis of the model).
This is demonstrated by the results in Tab. \ref{tab:results_emb_iam_time_testsetonly}, where we report the CPU time for the test set evaluation only.
This fact might assume more importance in particular applications, especially in those where the synthesis of the classifier can be effectively performed only once in off-line mode and the classification model is employed to process high-rate data streams in real-time \cite{6583538}.

Let us focus now on the structural complexity of the synthesized classification models.
The cardinality of the best-performing RSs are shown in Tab. \ref{tab:results_emb_iam_time}. It is possible to note that the cardinality are slightly bigger for those variants operating with MST-RE (especially in the first three datasets, i.e., L-L, L-M, and L-H).
From this fact we deduce that, when configuring the CBC procedure with the MST-RE estimator, the ODSE classifier, in order to obtain good results in terms of test set accuracy, requires a more complex model w.r.t. the variants involving the QRE estimator.
This behavior is however magnified by the setting of the objective function parameter $\eta$ adopted in our tests, which biases the ODSE system towards the recognition rate performance.
Notably, variants operating with the MMN develop considerable less costly classification models (see Tab. \ref{tab:results_emb_iam_rs} and \ref{tab:results_emb_iam_mmhb} for the details).
This particular aspect becomes very important in resource-constrained scenarios and/or when the input datasets are very big.
The considerable reductions of the RS size here achieved strengthen the fact that the entropy estimation operates adequately in the dissimilarity representation context.
\begin{table*}[tph!]\scriptsize
\caption{Test set classification accuracy results -- grayed lines denote novel results introduced in this paper. The ``-'' sign means that the result is not available to our knowledge.}
\begin{center}
\begin{tabular}{cccccccccccc}
\hline
\textbf{Classifier} & \multicolumn{8}{c}{\textbf{Dataset}} & \textbf{Rank} \\
\hline
& \textbf{L-L} & \textbf{L-M} & \textbf{L-H} & \textbf{AIDS} & \textbf{P} & \textbf{G} & \textbf{M} & \textbf{C-D} &  \\
\hline
\multicolumn{10}{c}{\textbf{Reference systems}} \\
\hline
RPS+TWEC+\textit{k}-NN, $k=1$ & 98.4 & 96.0 & 95.0 & 98.5 & 45.5 & 95.0 & 69.0 & 81.0 & 15 \\
\textit{k}-NN+TWEC, $k=1$ & 96.8 & 66.3 & 36.3 & 73.9 & 52.1 & 95.0 & 57.7 & 61.2 & 38 \\
RPS+TWEC+\textit{k}-NN, $k=3$ & 98.6 & 97.2 & 94.7 & 98.2 & 40.5 & 92.0 & 68.7 & 63.2 & 23 \\
\textit{k}-NN+TWEC, $k=3$ & 97.5 & 57.4 & 39.1 & 71.4 & 48.5 & 91.8 & 56.1 & 33.7 & 39 \\
RPS+TWEC+\textit{k}-NN, $k=5$ & 98.3 & 97.1 & 95.0 & 97.6 & 35.4 & 84.8 & 68.5 & 59.7 & 32 \\
\textit{k}-NN+TWEC, $k=5$ & 97.6 & 60.4 & 42.2 & 76.7 & 43.0 & 88.5 & 56.9 & 27.8 & 40 \\
RPS+TWEC+MMN & 98.0 & 96.0 & 93.6 & 97.4 & 49.5 & 95.0 & 66.0 & 68.4 & 28 \\
\hline
\multicolumn{10}{c}{\textbf{SOA systems}} \\
\hline
GMM+soft all+SVM \cite{Gibert20123072} & 99.7 & 93.0 & 87.8 & - & - & 99.0 & - & 98.1 & 12 \\
Fuzzy \textit{k}-means+soft all+SVM \cite{Gibert20123072} & 99.8 & 98.8 & 85.0 & - & - & 98.1 & - & 97.3 & 9 \\
sk+SVM \cite{riesen+bunke2009} & 99.7 & 85.9 & 79.1 & 97.4 & - & 94.4 & 55.4 & - & 30 \\
le+SVM \cite{riesen+bunke2009} & 99.3 & 95.9 & 92.5 & 98.3 & - & 96.8 & 74.3 & - & 7 \\
PCA+SVM \cite{riesen09Reducingdimensionalitydissimilarityspaceembeddinggraphkernels} & 92.7 & 81.1 & 73.3 & 98.2 & - & 92.9 & 75.9 & 93.6 & 26 \\
MDA+SVM \cite{riesen09Reducingdimensionalitydissimilarityspaceembeddinggraphkernels} & 89.8 & 68.5 & 60.5 & 95.4 & - & 91.8 & 62.4 & 88.2 & 37 \\
svm+SVM \cite{bunke10Improvingvectorspaceembeddinggraphsthroughfeatureselectionalgorithms} & 99.2 & 94.7 & 92.8 & 98.1 & 71.5 & 92.2 & 68.3 & - & 17 \\
svm+kPCA \cite{bunke10Improvingvectorspaceembeddinggraphsthroughfeatureselectionalgorithms} & 99.2 & 94.7 & 90.3 & 98.1 & 67.5 & 91.6 & 71.2 & - & 14 \\
lgq \cite{jain+srinivasan+tissen+obermayer2010} & 81.5 & - & - & - & - & 86.2 & - & - & 35 \\
bayes$_1$ \cite{jain+obermayer2011} & 80.4 & - & - & - & - & 80.3 & - & - & 36 \\
bayes$_2$ \cite{jain+obermayer2011} & 81.3 & - & - & - & - & 89.9 & - & - & 34 \\
FMGE+\textit{k}-NN \cite{Luqman:2013:FMG:2381462.2381562} & 97.1 & 75.7 & 66.5 & - & - & 97.5 & 69.1 & - & 31 \\
FMGE+SVM \cite{Luqman:2013:FMG:2381462.2381562} & 98.2 & 83.1 & 70.0 & - & - & 99.4 & 76.5 & - & 21 \\
d-sps-SVM \cite{ZareBorzeshi20131648} & 99.5 & 95.4 & 93.4 & 98.2 & 73.0 & 92.5 & 71.5 & - & 8 \\
GRALGv1 \cite{gralg_2012} & 98.2 & 75.6 & 69.6 & 99.7 & - & 97.7 & 73.0 & 94.0 & 10 \\
GRALGv2 \cite{gralg_2012} & 97.6 & 89.6 & 82.6 & 99.7 & 64.6 & 97.6 & 73.0 & 97.8 & 6 \\
\hline
\multicolumn{10}{c}{\textbf{Original ODSE}} \\
\hline
ODSE, $k=1$ \cite{odse}  & 98.6 & 96.8 & 96.2 & 99.6 & 61.0 & 96.2 & 73.4 & - & 1 \\
\hline
\multicolumn{10}{c}{\textbf{Improved ODSE with QRE}} \\
\hline
ODSE2v1, $k=1$ \cite{odse2_ijcnn_2013} & 99.0 & 97.0 & 96.1 & 99.1 & 61.2 & \textbf{98.1} & 68.2 & 78.1 & 4 \\
ODSE2v2, $k=1$ \cite{odse2_ijcnn_2013} & 98.7 & 97.1 & 95.4 & 99.5 & 51.9 & 95.4 & 68.1 & 77.2 & 5 \\
ODSE2v1, $k=3$ \cite{odse2_ijcnn_2013} & 99.0 & 97.2 & 96.1 & 99.3 & 41.4 & 90.2 & 68.7 & 64.3 & 13 \\
ODSE2v2, $k=3$ \cite{odse2_ijcnn_2013} & 98.8 & \textbf{97.4} & 95.1 & 99.4 & 31.4 & 38.0 & 69.4 & 59.0 & 24 \\
ODSE2v1, $k=5$ \cite{odse2_ijcnn_2013} & \textbf{99.1} & 96.8 & 95.2 & 99.0 & 38.9 & 85.4 & 69.0 & 58.6 & 27 \\
ODSE2v2, $k=5$ \cite{odse2_ijcnn_2013} & 98.7 & 97.0 & 95.6 & 99.4 & 31.3 & 82.5 & 70.0 & 54.0 & 25 \\
\rowcolor{lgray} ODSE2v1-MMN & 98.3 & 95.2 & 94.0 & 99.3 & 53.1 & 94.5 & 67.9 & 62.8 & 22 \\
\rowcolor{lgray} ODSE2v2-MMN & 97.8 & 95.6 & 93.6 & 99.6 & 48.7 & 94.8 & 68.2 & 59.2 & 29 \\
\hline
\multicolumn{10}{c}{\textbf{Improved ODSE with MST-RE}} \\
\hline
\rowcolor{lgray} ODSE2v1-MST, $k=1$ & 98.6 & 96.8 & \textbf{98.9} & 99.3 & 61.3 & 95.6 & 70.0 & 81.0 & 3 \\
\rowcolor{lgray} ODSE2v2-MST, $k=1$ & 98.4 & 97.1 & 96.0 & \textbf{99.7} & 51.0 & 94.1 & 71.6 & \textbf{82.0} & 2 \\
\rowcolor{lgray} ODSE2v1-MST, $k=3$ & 98.7 & 97.0 & 96.8 & 99.5 & 43.0 & 92.3 & 68.6 & 64.8 & 11 \\
\rowcolor{lgray} ODSE2v2-MST, $k=3$ & 98.8 & 96.9 & 96.0 & \textbf{99.7} & 35.0 & 91.0 & 69.4 & 60.0 & 16 \\
\rowcolor{lgray} ODSE2v1-MST, $k=5$ & 99.0 & 96.8 & 95.6 & 99.6 & 41.4 & 85.0 & 68.6 & 60.0 & 18 \\
\rowcolor{lgray} ODSE2v2-MST, $k=5$ & 98.8 & 97.0 & 95.5 & \textbf{99.7} & 32.9 & 83.3 & 70.0 & 54.0 & 19 \\
\rowcolor{lgray} ODSE2v1-MST-MMN & 97.9 & 95.4 & 93.6 & 99.3 & 49.9 & 95.0 & 68.3 & 62.6 & 20 \\
\rowcolor{lgray} ODSE2v2-MST-MMN & 97.9 & 95.1 & 91.8 & 99.2 & 48.5 & 94.8 & 67.1 & 59.0 & 33 \\
\hline
\end{tabular}
\end{center}
\label{tab:results_emb_iam}
\end{table*}
\begin{table*}[tph!]\scriptsize
\caption{Standard deviations of ODSE results shown in Tab. \ref{tab:results_emb_iam}.}
\begin{center}
\begin{tabular}{cccccccccccc}
\hline
\textbf{Classifier} & \multicolumn{8}{c}{\textbf{Dataset}} \\
\hline
& \textbf{L-L} & \textbf{L-M} & \textbf{L-H} & \textbf{AIDS} & \textbf{P} & \textbf{G} & \textbf{M} & \textbf{C-D} \\
\hline
ODSE \cite{odse} & 0.0256 & 1.2346 & 0.2423 & 0.0000 & 0.7356 & 0.4136 & 0.6586 & - \\
\hline
ODSE2v1, $k=1$ \cite{odse2_ijcnn_2013} & 0.0769 & 0.2309 & 0.1539 & 0.0000 & 2.6242 & 1.3350 & 0.5187 & 4.3863 \\
ODSE2v2, $k=1$ \cite{odse2_ijcnn_2013} & 0.0769 & 0.0769 & 0.4000 & 0.0000 & 0.2915 & 0.8021 & 0.5622 & 2.2654 \\
ODSE2v1, $k=3$ \cite{odse2_ijcnn_2013} & 0.0769 & 0.2309 & 0.2666 & 0.0000 & 1.0513 & 1.2236 & 0.0856 & 0.0577 \\
ODSE2v2, $k=3$ \cite{odse2_ijcnn_2013} & 0.0769 & 0.4618 & 5.0800 & 0.1924 & 1.1666 & 3.1540 & 0.0356 & 1.2361 \\
ODSE2v1, $k=5$ \cite{odse2_ijcnn_2013} & 0.5047 & 0.0769 & 0.9365 & 0.1924 & 0.5050 & 2.5585 & 0.3803 & 1.3279 \\
ODSE2v2, $k=5$ \cite{odse2_ijcnn_2013} & 0.1333 & 0.2309 & 0.0769 & 0.0000 & 2.7815 & 4.5220 & 1.2666 & 0.0026 \\
\rowcolor{lgray} ODSE2v1-MMN & 0.1520 & 0.3320 & 0.3932 & 0.1861 & 1.7740 & 0.7315 & 1.1300 & 1.0001 \\
\rowcolor{lgray} ODSE2v2-MMN & 0.2022 & 0.2022 & 0.7682 & 0.0000 & 2.7290 & 1.3584 & 1.4080 & 0.3896 \\
\hline
\rowcolor{lgray} ODSE2v1-MST, $k=1$ & 0.0730 & 0.0730 & 0.1115 & 0.2772 & 1.5500 & 0.1055 & 1.0786 & 0.4163 \\
\rowcolor{lgray} ODSE2v2-MST, $k=1$ & 0.0596 & 0.2231 & 0.0730 & 0.0000 & 1.1660 & 0.2943 & 0.9534 & 0.2146 \\
\rowcolor{lgray} ODSE2v1-MST, $k=3$ & 0.1192 & 0.1520 & 0.0942 & 0.6982 & 1.0940 & 0.0000 & 0.5926 & 1.7088 \\
\rowcolor{lgray} ODSE2v2-MST, $k=3$ & 0.1460 & 0.2022 & 0.0730 & 0.0000 & 0.0000 & 0.1112 & 0.2365 & 0.5655 \\
\rowcolor{lgray} ODSE2v1-MST, $k=5$ & 0.1115 & 0.0942 & 0.2190 & 0.0596 & 0.4748 & 0.0000 & 0.0547 & 1.2356 \\
\rowcolor{lgray} ODSE2v2-MST, $k=5$ & 0.0730 & 0.0596 & 0.9933 & 0.0000 & 0.0000 & 0.1112 & 1.0023 & 0.9563 \\
\rowcolor{lgray} ODSE2v1-MST-MMN & 0.1115 & 0.4216 & 0.7624 & 0.3217 & 2.5735 & 0.3067 & 0.7926 & 0.9899 \\
\rowcolor{lgray} ODSE2v2-MST-MMN & 0.0596 & 0.7636 & 0.7477 & 0.0000 & 2.7290 & 0.5828 & 0.8911 & 1.2020 \\
\hline
\end{tabular}
\end{center}
\label{tab:results_emb_iam_sd}
\end{table*}
\begin{table*}[tph!]\scriptsize
\caption{Average serial CPU time in minutes (and speed-up factor w.r.t. the original ODSE system) considering ODSE model synthesis and test set evaluation. In the \textit{k}-NN case, we report the results with $k=1$ only.}
\begin{center}
\begin{tabular}{cccccccccccc}
\hline
\textbf{Classifier} & \multicolumn{8}{c}{\textbf{Dataset}} \\
\hline
& \textbf{L-L} & \textbf{L-M} & \textbf{L-H} & \textbf{AIDS} & \textbf{P} & \textbf{G} & \textbf{M} & \textbf{C-D} \\
\hline
ODSE \cite{odse} & 63274 & 52285 & 28938 & 394 & 8460 & 601 & 43060 & - \\
\hline
ODSE2v1 \cite{odse2_ijcnn_2013} & 284 (222) & 329 (158) & 328 (88) & 38 (10) & 3187 (3) & 210 (3) & 3494 (12) & 2724 \\
ODSE2v2 \cite{odse2_ijcnn_2013} & 126 (502) & 268 (195) & 183 (158) & 110 (3) & 1683 (5) & 96 (6) & 10326 (4) & 8444 \\
\rowcolor{lgray} ODSE2v1-MMN & 129 (490) & 284 (184) & 263 (110) & 17 (23) & 3638 (2) & 170 (4) & 8837 (5) & 5320 \\
\rowcolor{lgray} ODSE2v2-MMN & 195 (324) & 422 (124) & 183 (158) & 86 (5) & 1444 (6) & 77 (8) & 28511 (2) & 20301 \\
\hline
\rowcolor{lgray} ODSE2v1-MST & 213 (297) & 231 (226) & 225 (129) & 18 (22) & 3860 (2) & 168 (4) & 2563 (17) & 3261 \\
\rowcolor{lgray} ODSE2v2-MST & 145 (463) & 160 (327) & 107 (270) & 93 (4) & 2075 (4) & 74 (8) & 7675 (6) & 10092 \\
\rowcolor{lgray} ODSE2v1-MST-MMN & 201 (315) & 249 (210) & 205 (141) & 15 (26) & 3450 (2) & 155 (4) & 5496 (8) & 7135 \\
\rowcolor{lgray} ODSE2v2-MST-MMN & 117 (541) & 176 (292) & 118 (245) & 83 (5) & 1380 (6) & 75 (8) & 28007 (2) & 16599 \\
\hline
\end{tabular}
\end{center}
\label{tab:results_emb_iam_time}
\end{table*}
\begin{table*}[tph!]\scriptsize
\caption{Average serial CPU time in seconds for test set evaluation only. For simplicity, we report the results of only one system variant operating in the DS with the \textit{k}-NN classifier and only one with the MMN.}
\begin{center}
\begin{tabular}{cccccccccccc}
\hline
\textbf{Class. Sys.} & \multicolumn{8}{c}{\textbf{Datasets}} \\
\hline
& \textbf{L-L} & \textbf{L-M} & \textbf{L-H} & \textbf{AIDS} & \textbf{P} & \textbf{G} & \textbf{M} & \textbf{C-D} \\
\hline
\rowcolor{lgray} ODSE2v1-MST, $k=1$ & 0.740 & 0.740 & 0.740 & 0.130 & 0.020 & 0.060 & 9.020 & 9.700 \\
\hline
\rowcolor{lgray} ODSE2v1-MST-MMN & 0.105 & 0.105 & 0.105 & 0.005 & 0.014 & 0.045 & 6.600 & 5.250 \\
\hline
\end{tabular}
\end{center}
\label{tab:results_emb_iam_time_testsetonly}
\end{table*}
\begin{table*}[tph!]\scriptsize
\caption{Average cardinality of the best-performing RS. In the \textit{k}-NN case, we report the results with $k=1$ only since results with $k=3$ and $k=5$ are similar.}
\begin{center}
\begin{tabular}{cccccccccccc}
\hline
\textbf{Classifier} & \multicolumn{8}{c}{\textbf{Dataset}} \\
\hline
& \textbf{L-L} & \textbf{L-M} & \textbf{L-H} & \textbf{AIDS} & \textbf{P} & \textbf{G} & \textbf{M} & \textbf{C-D} \\
\hline
ODSE \cite{odse} & 435 & 750 & 750 & 250 & 200 & 283 & 1500 & - \\
\hline
ODSE2v1 \cite{odse2_ijcnn_2013} & 146 & 449 & 449 & 8 & 197 & 283 & 760 & 615 \\
ODSE2v2 \cite{odse2_ijcnn_2013} & 183 & 431 & 338 & 7 & 82 & 126 & 801 & 770 \\
\rowcolor{lgray} ODSE2v1-MM & 136 & 192 & 144 & 6 & 190 & 163 & 563 & 555 \\
\rowcolor{lgray} ODSE2v2-MM & 197 & 546 & 80 & 2 & 93 & 115 & 815 & 740 \\
\hline
\rowcolor{lgray} ODSE2v1-MST & 597 & 595 & 597 & 6 & 198 & 283 & 687 & 618 \\
\rowcolor{lgray} ODSE2v2-MST & 551 & 574 & 447 & 61 & 122 & 129 & 813 & 775 \\
\rowcolor{lgray} ODSE2v1-MST-MMN & 600 & 606 & 500 & 5 & 190 & 184 & 424 & 549 \\
\rowcolor{lgray} ODSE2v2-MST-MMN & 550 & 580 & 411 & 61 & 93 & 115 & 456 & 733 \\
\hline
\end{tabular}
\end{center}
\label{tab:results_emb_iam_rs}
\end{table*}
\begin{table*}[tph!]\scriptsize
\caption{Average number of hyperboxes generated by the MMN. The number of hyperboxes can be used also as a complexity indicator of the model synthesized by the MMN on the DS. Such values should be taken into account considering also the dataset characteristics of Tab. \ref{tab:iam_details} and the computed average representation set sizes in Tab. \ref{tab:results_emb_iam_rs}.}
\begin{center}
\begin{tabular}{cccccccccccc}
\hline
\textbf{Classifier} & \multicolumn{8}{c}{\textbf{Dataset}} \\
\hline
& \textbf{L-L} & \textbf{L-M} & \textbf{L-H} & \textbf{AIDS} & \textbf{P} & \textbf{G} & \textbf{M} & \textbf{C-D} \\
\hline
\rowcolor{lgray} ODSE2v1-MMN & 15 & 39 & 34 & 5 & 43 & 27 & 164 & 357 \\
\rowcolor{lgray} ODSE2v2-MMN & 15 & 28 & 41 & 4 & 48 & 28 & 159 & 368 \\
\hline
\rowcolor{lgray} ODSE2v1-MST-MMN & 15 & 27 & 38 & 3 & 48 & 28 & 168 & 348 \\
\rowcolor{lgray} ODSE2v2-MST-MMN & 15 & 27 & 34 & 4 & 43 & 27 & 175 & 365 \\
\hline
\end{tabular}
\end{center}
\label{tab:results_emb_iam_mmhb}
\end{table*}

\section{Conclusions and future directions}
\label{sec:conclusions}

In this paper, we have presented different variants of the improved ODSE graph classification system. All the discussed variants are based on the characterization of the informativeness of the DM through the estimation of the $\alpha$-order R\'{e}nyi entropy.
The first adopted estimator computes the QRE by means of a kernel-based density estimator, while the second one uses the length of an entropic MST.
The improved ODSE system has been designed by providing different strategies for the initialization, compression, as well as for the expansion operation of the RS. In particular, we conceived a fast CBC scheme, which allowed us to directly control the compression level of the data through the explicit setting of the cluster radius parameter.
We provided formal proofs for the two estimation techniques. These proofs enabled us to determine the value of the cluster radius analytically, according to the ODSE model optimization procedure.
We have studied also the asymptotic worst-case efficiency of the CBC scheme implemented by means of a sequential cluster generation rule (BSAS).

Experimental evaluations and comparisons with several state-of-the-art systems have been performed on well-known benchmarking datasets of labeled graphs (IAM database).
We used two different feature-based classifiers operating in the DS: the \textit{k}-NN classifier equipped with the Euclidean distance and a neurofuzzy MMN trained with the ARC algorithm.
Overall, the variants adopting the MST-based estimator resulted to be faster but less parsimonious for what concerns the synthesized ODSE model (i.e., the cardinality of the best-performing RS was larger). The use of the \textit{k}-NN rule (with $k=1$) yielded slightly better test set accuracy results w.r.t. the MMN, while however in the latter case we have observed important differences in term of (serial) CPU computing time, especially on the test set processing stage.
The test set classification accuracy results confirmed the effectiveness of the ODSE classifier w.r.t. state-of-the-art standards. Moreover, the significative CPU time improvements w.r.t. the original ODSE version, and the highly parallelizable global optimization scheme based on a genetic algorithm, bring the ODSE graph classifier one step closer towards the applicability to bigger labeled graphs and larger datasets.

The vector representation of the input graphs have been obtained directly using the rows of the dissimilarity matrix. Such a choice, while it is known to be effective, has been mainly dictated by the computing time requirements of the system. It is worth analyzing the performance of ODSE also when the embedding space is obtained by a (non)linear embedding of the (corrected) pairwise dissimilarity values \cite{wilson2014spherical}.
Future experiments include testing other core IGM procedures, different $\alpha$-order R\'{e}nyi entropy estimators, and additional feature-based classifiers.

\appendix
\section{Proof of Theorem \ref{th:thmSequenceOrdering}}
\label{sec:appendixA}
\begin{proof}
We focus on the worst-case scenario for $\xi$, giving thus a lower bound for the efficiency (\ref{eq:efficiency}).
Let $s[i]=\underline{\mathbf{x}}_i$ denote the \textit{i}-th element of the sequence $s$, i.e., the \textit{i}-th dissimilarity vector corresponding to the prototype graph $R_i\in\mathcal{R}$.
Let $s^{*}$ be the best ordering for $s$, i.e.,
\begin{equation}
s^{*} = \argmax_{s_i\in\Omega(s)} \rho(s_i).
\end{equation}

Let us assume the case in which the Euclidean distance among any pair of vectors in $s$ is given by
\begin{equation}
\label{eq:worst_case_distance}
d_2(s[i], s[j])=|i-j|\theta, \ 1 \leq i,j \leq n,
\end{equation}
where $\theta$ is the adopted cluster radius during the ODSE compression.
It is easy to understand that this is the worst-case scenario for the compression purpose in the sequential clustering setting. 
In fact, each vector $\underline{\mathbf{x}}_i$ in the sequence $s$ has a distance with its predecessor/successor equal to the maximum cluster radius $\theta$.
As a consequence, there is still a possibility to compress the vectors, but it is however strictly dependent on the specific ordering of $s$.

First of all, it is important to note that, due to the distances assumed in (\ref{eq:worst_case_distance}), only three elements of $s$ can be contained into a single cluster. In fact, any three consecutive elements of the sequence $s$ would form a cluster with a diameter equal to $2\theta$. Therefore, considering the sequential rule shown in Algorithm \ref{alg:bsas}, and setting $Q=n$, the best possible ordering $s^{*}$ is the one that preserves a distance equal to $\theta$ for any two adjacent elements of $s$, achieving a compression ratio of:
\begin{equation}
\label{eq:lower_bound}
\rho^{*}(s) = n / \lceil n/3 \rceil .
\end{equation}

The worst possible ordering, instead, yields $n / \lceil n/2 \rceil$, which can be achieved (for instance assuming $n$ odd) when considering the following ordering $s_i$ w.r.t. the optimal $s^{*}$:
\begin{equation}
\label{eq:worst_ordering}
s_i[j] = s^{*}[(2j \mbox{ mod } n)+1], j=1, 2, ..., n.
\end{equation}

In this case, Algorithm \ref{alg:bsas} would generate exactly 
\begin{equation}
\label{eq:upper_bound}
\lceil n/2 \rceil
\end{equation}
clusters, corresponding to the first $\lceil n/2 \rceil$ elements of the sequence $s_i$, since every pair of consecutive elements in $s_i$ is at a distance of exactly $2\theta$.
Therefore, $\lceil n/2 \rceil$ is the maximum number of clusters that can be generated by considering the distances assumed in (\ref{eq:worst_case_distance}).
Combining Eq. \ref{eq:lower_bound} and \ref{eq:upper_bound}, we obtain for a given $s$,
\begin{equation}
\label{eq:effective_vs_optimal_cr}
n / \lceil n/2 \rceil \leq \hat{\rho}(s) \leq \rho^{*}(s) = n / \lceil n/3 \rceil,
\end{equation}
which allows us to claim that the worst-case efficiency of the ODSE compression varies according to the following ratio:
\begin{equation}
\label{eq:odse_compression_eff}
\hat{\rho}(s) / \rho^{*}(s) = \frac{n}{\lceil n/2 \rceil}\times \frac{\lceil n/3 \rceil}{n} = \frac{\lceil n/3 \rceil}{\lceil n/2 \rceil} .
\end{equation}

Taking the limit for $n\rightarrow\infty$ in Eq. \ref{eq:odse_compression_eff} gives us the claim.
\end{proof}

\section{Proof of Theorem \ref{th:thmThetaMST}}
\label{sec:appendixB}

\begin{proof}
Let us focus the analysis on a single cluster $\mathcal{B}\in P(\theta; \tau_c, \gamma)$, containing $k=|\mathcal{B}|$ prototypes within a training set of $n$ graphs.
Let us remind that the cluster radius and diameter are, respectively, $\theta$ and $2\theta$ in the spherical cluster case.
Therefore, we can obtain an upper bound for the MST length factor (\ref{eq:l_mst}), considering that (all) the corresponding MST, $T$, of the complete graph generated from the $k$ measurements has $k-1$ edges with weights equal to $2\theta$. Specifically,
\begin{equation}
\label{eq:bound_l_mst}
L_{\gamma}(\theta) = \displaystyle\sum_{e_{ij}\in T} |e_{ij}|^{\gamma} = (k-1)\times (2\theta)^{\gamma}.
\end{equation}

In the following, we evaluate $\beta(L_{\gamma}(\theta), n)$ exactly as defined in Eq. \ref{eq:beta_approx}, considering $n$ dimensions -- note that $\beta(L_{\gamma}(\theta), n)$ is shortened as $\beta$.
Eq. \ref{eq:bound_l_mst} allows us to derive the following upper bound for the MST-based entropy estimator (\ref{eq:rentropy_mst}):
\begin{align}
\nonumber\hat{H}_{\alpha}(\underline{\mathbf{D}}_{k}) &= \frac{n}{\gamma}\left[ \ln\left(\frac{L_{\gamma}(\underline{\mathbf{D}}_{k})}{k^{\alpha}}\right) - \ln\left(\beta(L_{\gamma}, n)\right) \right] \\
\nonumber& \leq \frac{n}{\gamma}\left[ \ln\left(\frac{L_{\gamma}(\theta)}{k^{\alpha}}\right) - \ln(\beta) \right] \\
\nonumber& = \frac{n}{\gamma}\left[ \ln\left(\frac{(k-1)\times(2\theta)^{\gamma}}{k^{\alpha}}\right) - \ln(\beta) \right] \\
\label{eq:mark1}
& = \frac{n}{\gamma}\left[ \ln(k-1) + \gamma\ln(2\theta)  - \ln(k^{\alpha}) - \ln(\beta) \right] .
\end{align}

However, the entropy estimator shown in Eq. \ref{eq:rentropy_mst} does not yield normalized values (e.g., in $[0, 1]$).
We can normalize the estimations by considering the following factor:
\begin{equation}
\label{eq:normalization_f}
\iota = \frac{n}{\gamma}\left[ \ln(k-1) + \gamma\ln(\Delta\sqrt{n}) - \ln(k^{\alpha}) - \ln(\beta) \right].
\end{equation}

The quantity $\Delta\sqrt{n}$ is the maximum distance in an Euclidean $\Delta$-hypercube of $n$-dimensions; $\Delta$ is the input data extent, which is 2 in our case. Eq. \ref{eq:normalization_f} is a maximizer of Eq. \ref{eq:rentropy_mst} since the logarithm is a monotonically increasing function and the other relevant factors in the expression remain constant changing the input distribution.
Instead, the MST length achieves its maximum value only in the specific case when all $k$ points are at a distance equal to $2\sqrt{n}$.
Therefore, by normalizing Eq. \ref{eq:mark1} using (\ref{eq:normalization_f}), we obtain:
\begin{equation}
\frac{\ln(k-1) + \gamma\ln(2\theta) - \ln(k^{\alpha}) - \ln(\beta)}{\ln(k-1) + \gamma\ln(2\sqrt{n}) - \ln(k^{\alpha}) - \ln(\beta)} \in [0, 1] .
\end{equation}

Rewriting the expression in terms of the ODSE compression rule (\ref{eq:compression_rule}), we have:
\begin{equation}
\label{eq:compressibility_norm}
\frac{\hat{H}_{\alpha}(\underline{\mathbf{D}}_{k})}{\iota} \leq \frac{\ln(k-1) + \gamma\ln(2\theta) - \ln(k^{\alpha}) - \ln(\beta)}{\ln(k-1) + \gamma\ln(2\sqrt{n}) - \ln(k^{\alpha}) - \ln(\beta)} \leq \tau_c .
\end{equation}

Solving for $\theta$, the right-hand side of (\ref{eq:compressibility_norm}) can be manipulated as follows:
\begin{align}
\nonumber\gamma\ln(2\theta) \leq& \tau_c \left[ \ln(k-1) + \gamma\ln(2\sqrt{n}) - \ln(k^{\alpha}) - \ln(\beta) \right] \\
\nonumber &- \ln(k-1) + \ln(k^{\alpha}) + \ln(\beta) ;\\
\nonumber\ln(2\theta) \leq& \frac{\tau_c}{\gamma} \left[ \ln(k-1) + \gamma\ln(2\sqrt{n}) - \ln(k^{\alpha}) -\ln(\beta) \right] \\
\nonumber &+ \frac{1}{\gamma}\left[ -\ln(k-1) + \ln(k^{\alpha}) + \ln(\beta) \right];\\
\nonumber\theta \leq& \frac{1}{2} \exp\left( \frac{\tau_c}{\gamma} \left[ \ln(k-1) + \gamma\ln(2\sqrt{n}) - \ln(k^{\alpha}) -\ln(\beta) \right] \right) \\
\nonumber &\times \exp\left( \frac{1}{\gamma} \left[ -\ln(k-1) + \ln(k^{\alpha}) + \ln(\beta) \right] \right) ; \\
\nonumber\theta \leq& \frac{1}{2} \left[ \exp\left( \ln(k-1) + \gamma\ln(2\sqrt{n}) - \ln(k^{\alpha}) -\ln(\beta) \right) \right]^{\frac{\tau_c}{\gamma}} \\
\nonumber &\times \left[ \exp \left( -\ln(k-1) + \ln(k^{\alpha}) + \ln(\beta) \right) \right]^{\frac{1}{\gamma}} ; \\
\nonumber\theta \leq& \frac{1}{2} \left[ (k-1)  2^{\gamma}  n^{\frac{\gamma}{2}}  k^{-\alpha}  \beta^{-1} \right]^{\frac{\tau_c}{\gamma}}  \left[ (k-1)^{-1}  k^{\alpha}  \beta \right]^{\frac{1}{\gamma}} ; \\
\nonumber\theta \leq& \frac{1}{2}  (k-1)^{\frac{\tau_c}{\gamma}}  2^{\tau_c}  n^{\frac{\tau_c}{2}}  k^{\frac{-\alpha\tau_c}{\gamma}}  \beta^{-\frac{\tau_c}{\gamma}}  (k-1)^{-\frac{1}{\gamma}}  k^{\frac{\alpha}{\gamma}}  \beta^{\frac{1}{\gamma}} ; \\
\label{eq:mark2}
\theta \leq& (k-1)^{\frac{\tau_c-1}{\gamma}}  2^{\tau_c-1} n^{\frac{\tau_c}{2}}  k^{\frac{\alpha(-\tau_c+1)}{\gamma}} \beta^{\frac{-\tau_c+1}{\gamma}} .
\end{align}

Considering that $\tau_c-1\leq 0$ and $(-\tau_c+1)\in[0, 1]$ hold for any $\tau_c\in[0, 1]$, we rewrite Eq. \ref{eq:mark2} accordingly as follows:
\begin{align}
\theta &\leq 2^{\tau_c-1} n^{\frac{\tau_c}{2}} \beta^{\frac{-\tau_c+1}{\gamma}} \frac{k^{\frac{\alpha(-\tau_c+1)}{\gamma}}}{(k-1)^{\frac{-\tau_c+1}{\gamma}}} ; \\
\label{eq:alpha_power}
\theta &\leq 2^{\tau_c-1} n^{\frac{\tau_c}{2}} \beta^{\frac{-\tau_c+1}{\gamma}} \left(\frac{k^{\alpha}}{(k-1)}\right)^{\frac{-\tau_c+1}{\gamma}}.
\end{align}

The right-hand side of Eq. \ref{eq:alpha_power} can be further simplified in:
\begin{equation}
\theta \leq  2^{\tau_c-1} n^{\frac{\tau_c}{2}} \beta^{\frac{-\tau_c+1}{\gamma}} c(\gamma) ,
\end{equation}
where the $c(\gamma)$ function has the following bounds:
\begin{equation}
\label{eq:constant}
0 \leq c(\gamma) \leq \left(\frac{k^{\alpha}}{k-1}\right)^{\frac{-\tau_c+1}{\gamma}} .
\end{equation}

In fact, provided that $\alpha\in(0, 1)$ and $k\in\mathbb{N}$ hold, with $k\geq2$ (there is no need to compress singleton clusters), we have:
\begin{equation}
\label{eq:cases_constant}
\begin{cases}
k^{\frac{\alpha(-\tau_c+1)}{\gamma}} (k-1)^{\frac{\tau_c-1}{\gamma}} = 0 &\mbox{ if } k\rightarrow\infty, \\
k^{\frac{\alpha(-\tau_c+1)}{\gamma}} (k-1)^{\frac{\tau_c-1}{\gamma}} = \left(\frac{k^{\alpha}}{k-1}\right)^{\frac{-\tau_c+1}{\gamma}} &\mbox{ otherwise}.
\end{cases}
\end{equation}

Note that $c(\gamma)$ depends also on $\alpha$, which, however, in turn depends on $\gamma$ (\ref{eq:alpha}); as a convention we express $c(\gamma)$ as a function of the $\gamma$ parameter only.
Eq. \ref{eq:cases_constant} evaluates to $2^{\frac{\alpha}{\gamma}}$ when $k=2$ and $\tau_c=0$, providing hence the upper bound for $c(\cdot)$.
\end{proof}

\bibliographystyle{abbrvnat}
\bibliography{/home/lorenzo/University/Research/Publications/Bibliography.bib}

\begin{thebibliography}{54}
\providecommand{\natexlab}[1]{#1}
\providecommand{\url}[1]{\texttt{#1}}
\expandafter\ifx\csname urlstyle\endcsname\relax
  \providecommand{\doi}[1]{doi: #1}\else
  \providecommand{\doi}{doi: \begingroup \urlstyle{rm}\Url}\fi

\bibitem[Bai and Hancock(2013)]{Bai_Hancock__2012}
L.~Bai and E.~R. Hancock.
\newblock {Graph Kernels from the {J}ensen-{S}hannon Divergence}.
\newblock \emph{Journal of Mathematical Imaging and Vision}, 47\penalty0
  (1-2):\penalty0 60--69, 2013.
\newblock \doi{10.1007/s10851-012-0383-6}.

\bibitem[Bai et~al.(2015)Bai, Rossi, Torsello, and Hancock]{bai2014quantum}
L.~Bai, L.~Rossi, A.~Torsello, and E.~R. Hancock.
\newblock A quantum {J}ensen--{S}hannon graph kernel for unattributed graphs.
\newblock \emph{Pattern Recognition}, 48\penalty0 (2):\penalty0 344--355, 2015.
\newblock ISSN 0031-3203.
\newblock \doi{10.1016/j.patcog.2014.03.028}.

\bibitem[Bengoetxea et~al.(2002)Bengoetxea, Larra{\~n}aga, Bloch, Perchant, and
  Boeres]{Bengoetxea20022867}
E.~Bengoetxea, P.~Larra{\~n}aga, I.~Bloch, A.~Perchant, and C.~Boeres.
\newblock {Inexact graph matching by means of estimation of distribution
  algorithms}.
\newblock \emph{Pattern Recognition}, 35\penalty0 (12):\penalty0 2867--2880,
  2002.
\newblock ISSN 0031-3203.
\newblock \doi{10.1016/S0031-3203(01)00232-1}.

\bibitem[Bianchi et~al.(2013)Bianchi, Livi, and Rizzi]{dyngraph_ijcnn_2013}
F.~M. Bianchi, L.~Livi, and A.~Rizzi.
\newblock Matching of time-varying labeled graphs.
\newblock In \emph{{Proceedings of the IEEE International Joint Conference on
  Neural Networks}}, pages 1660--1667, Aug 2013.
\newblock ISBN 978-1-4673-6129-3.
\newblock \doi{10.1109/IJCNN.2013.6706939}.

\bibitem[Bianchi et~al.(2014)Bianchi, Livi, Rizzi, and Sadeghian]{gralg_2012}
F.~M. Bianchi, L.~Livi, A.~Rizzi, and A.~Sadeghian.
\newblock A {G}ranular {C}omputing approach to the design of optimized graph
  classification systems.
\newblock \emph{Soft Computing}, 18\penalty0 (2):\penalty0 393--412, 2014.
\newblock ISSN 1432-7643.
\newblock \doi{10.1007/s00500-013-1065-z}.

\bibitem[Bonev et~al.(2008)Bonev, Escolano, and Cazorla]{bonev__2008}
B.~Bonev, F.~Escolano, and M.~Cazorla.
\newblock {Feature selection, mutual information, and the classification of
  high-dimensional patterns}.
\newblock \emph{Pattern Analysis and Applications}, 11\penalty0 (3-4):\penalty0
  309--319, 2008.
\newblock ISSN 1433-7541.
\newblock \doi{10.1007/s10044-008-0107-0}.

\bibitem[Borzeshi et~al.(2013)Borzeshi, Piccardi, Riesen, and
  Bunke]{ZareBorzeshi20131648}
E.~Z. Borzeshi, M.~Piccardi, K.~Riesen, and H.~Bunke.
\newblock {Discriminative prototype selection methods for graph embedding}.
\newblock \emph{Pattern Recognition}, 46\penalty0 (6):\penalty0 1648--1657,
  2013.
\newblock ISSN 0031-3203.
\newblock \doi{10.1016/j.patcog.2012.11.020}.

\bibitem[Bul\`{o} and Pelillo(2013)]{10.1109/TPAMI.2012.226}
S.~R. Bul\`{o} and M.~Pelillo.
\newblock A game-theoretic approach to hypergraph clustering.
\newblock \emph{IEEE Transactions on Pattern Analysis and Machine
  Intelligence}, 35:\penalty0 1312--1327, Jun. 2013.
\newblock ISSN 0162-8828.
\newblock \doi{10.1109/TPAMI.2012.226}.

\bibitem[Bunke and
  Riesen(2011)]{bunke10Improvingvectorspaceembeddinggraphsthroughfeatureselectionalgorithms}
H.~Bunke and K.~Riesen.
\newblock {Improving vector space embedding of graphs through feature selection
  algorithms}.
\newblock \emph{Pattern Recognition}, 44:\penalty0 1928--1940, 2011.
\newblock \doi{10.1016/j.patcog.2010.05.016}.

\bibitem[Bunke and Riesen(2012)]{Bunke2012811}
H.~Bunke and K.~Riesen.
\newblock {Towards the unification of structural and statistical pattern
  recognition}.
\newblock \emph{Pattern Recognition Letters}, 33\penalty0 (7):\penalty0
  811--825, 2012.
\newblock ISSN 0167-8655.
\newblock \doi{10.1016/j.patrec.2011.04.017}.

\bibitem[{Cesar Jr} et~al.(2005){Cesar Jr}, Bengoetxea, Bloch, and
  Larra{\~n}aga]{CesarJr20052099}
R.~M. {Cesar Jr}, E.~Bengoetxea, I.~Bloch, and P.~Larra{\~n}aga.
\newblock {Inexact graph matching for model-based recognition: Evaluation and
  comparison of optimization algorithms}.
\newblock \emph{Pattern Recognition}, 38\penalty0 (11):\penalty0 2099--2113,
  2005.
\newblock ISSN 0031-3203.
\newblock \doi{10.1016/j.patcog.2005.05.007}.

\bibitem[Chen(2014)]{chen2014type}
L.~Chen.
\newblock {EM}-type method for measuring graph dissimilarity.
\newblock \emph{International Journal of Machine Learning and Cybernetics},
  5\penalty0 (4):\penalty0 625--633, 2014.
\newblock \doi{10.1007/s13042-013-0210-4}.

\bibitem[{Del Vescovo} et~al.(2014){Del Vescovo}, Livi, {Frattale Mascioli},
  and Rizzi]{delvescovo+livi+rizzi+frattalemascioli2011}
G.~{Del Vescovo}, L.~Livi, F.~M. {Frattale Mascioli}, and A.~Rizzi.
\newblock On the problem of modeling structured data with the {MinSOD}
  representative.
\newblock \emph{International Journal of Computer Theory and Engineering},
  6\penalty0 (1):\penalty0 9--14, 2014.
\newblock ISSN 1793-8201.
\newblock \doi{10.7763/IJCTE.2014.V6.827}.

\bibitem[{Di Paola} et~al.(2012){Di Paola}, {De Ruvo}, Paci, Santoni, and
  Giuliani]{doi:10.1021/cr3002356}
L.~{Di Paola}, M.~{De Ruvo}, P.~Paci, D.~Santoni, and A.~Giuliani.
\newblock Protein contact networks: {A}n emerging paradigm in {C}hemistry.
\newblock \emph{Chemical Reviews}, 113\penalty0 (3):\penalty0 1598--1613, 2012.
\newblock \doi{10.1021/cr3002356}.

\bibitem[D{\"o}rfler and Bullo(2013)]{dorfler+bullo2011}
F.~D{\"o}rfler and F.~Bullo.
\newblock Kron reduction of graphs with applications to electrical networks.
\newblock \emph{IEEE Transactions on Circuits and Systems}, 60\penalty0
  (1):\penalty0 150--163, Jan. 2013.
\newblock \doi{10.1109/TCSI.2012.2215780}.

\bibitem[Emmert-Streib et~al.(2016)Emmert-Streib, Dehmer, and
  Shi]{emmert2016fifty}
F.~Emmert-Streib, M.~Dehmer, and Y.~Shi.
\newblock Fifty years of graph matching, network alignment and network
  comparison.
\newblock \emph{Information Sciences}, 346:\penalty0 180--197, 2016.
\newblock \doi{10.1016/j.ins.2016.01.074}.

\bibitem[Escolano et~al.(2011)Escolano, Bonev, and Lozano]{escolano_2011}
F.~Escolano, B.~Bonev, and M.~Lozano.
\newblock {Information-Geometric Graph Indexing from Bags of Partial Node
  Coverages}.
\newblock In X.~Jiang, M.~Ferrer, and A.~Torsello, editors, \emph{{Graph-Based
  Representations in Pattern Recognition}}, volume 6658 of \emph{{LNCS}}, pages
  52--61. Springer Berlin / Heidelberg, 2011.
\newblock ISBN 978-3-642-20843-0.
\newblock 10.1007/978-3-642-20844-7\_6.

\bibitem[Escolano et~al.(2013)Escolano, Hancock, Liu, and
  Lozano]{it_diss_graphs__Escolano_2013}
F.~Escolano, E.~R. Hancock, M.~Liu, and M.~Lozano.
\newblock {Information-Theoretic Dissimilarities for Graphs}.
\newblock In E.~Hancock and M.~Pelillo, editors, \emph{{Similarity-Based
  Pattern Recognition}}, volume 7953, pages 90--105. Springer Berlin,
  Heidelberg, 2013.
\newblock ISBN 978-3-642-39139-2.
\newblock \doi{10.1007/978-3-642-39140-8_6}.

\bibitem[Filippone et~al.(2008)Filippone, Camastra, Masulli, and
  Rovetta]{FilipponePR08}
M.~Filippone, F.~Camastra, F.~Masulli, and S.~Rovetta.
\newblock {A survey of kernel and spectral methods for clustering}.
\newblock \emph{Pattern Recognition}, 41\penalty0 (1):\penalty0 176--190, Jan.
  2008.
\newblock \doi{10.1016/j.patcog.2010.08.001}.

\bibitem[Gibert et~al.(2012)Gibert, Valveny, and Bunke]{Gibert20123072}
J.~Gibert, E.~Valveny, and H.~Bunke.
\newblock {Graph embedding in vector spaces by node attribute statistics}.
\newblock \emph{Pattern Recognition}, 45\penalty0 (9):\penalty0 3072--3083,
  2012.
\newblock ISSN 0031-3203.
\newblock \doi{10.1016/j.patcog.2012.01.009}.

\bibitem[Hancock and Wilson(2012)]{Hancock2012833}
E.~R. Hancock and R.~C. Wilson.
\newblock {Pattern analysis with graphs: Parallel work at Bern and York}.
\newblock \emph{Pattern Recognition Letters}, 33\penalty0 (7):\penalty0
  833--841, 2012.
\newblock ISSN 0167-8655.
\newblock \doi{10.1016/j.patrec.2011.08.012}.

\bibitem[{Hero III} and Michel(1999)]{Hero_Asympt__1999}
A.~O. {Hero III} and O.~J.~J. Michel.
\newblock {Asymptotic theory of greedy approximations to minimal k-point random
  graphs}.
\newblock \emph{IEEE Transactions on Information Theory}, 45:\penalty0
  1921--1938, Sep. 1999.
\newblock ISSN 0018-9448.
\newblock \doi{10.1109/18.782114}.

\bibitem[Jain(2016{\natexlab{a}})]{jain2016geometry}
B.~J. Jain.
\newblock On the geometry of graph spaces.
\newblock \emph{Discrete Applied Mathematics}, 214:\penalty0 126--144,
  2016{\natexlab{a}}.
\newblock \doi{10.1016/j.dam.2016.06.027}.

\bibitem[Jain(2016{\natexlab{b}})]{jain2016statistical}
B.~J. Jain.
\newblock Statistical graph space analysis.
\newblock \emph{Pattern Recognition}, 60:\penalty0 802--812,
  2016{\natexlab{b}}.
\newblock \doi{10.1016/j.patcog.2016.06.023}.

\bibitem[Jain and Obermayer(2011)]{jain+obermayer2011}
B.~J. Jain and K.~Obermayer.
\newblock {Maximum Likelihood for {G}aussians on Graphs}.
\newblock In X.~Jiang, M.~Ferrer, and A.~Torsello, editors, \emph{{Graph-Based
  Representations in Pattern Recognition}}, volume 6658, pages 62--71. Springer
  Berlin, Heidelberg, 2011.
\newblock \doi{10.1007/978-3-642-20844-7_7}.

\bibitem[Jain et~al.(2010)Jain, Srinivasan, Tissen, and
  Obermayer]{jain+srinivasan+tissen+obermayer2010}
B.~J. Jain, S.~D. Srinivasan, A.~Tissen, and K.~Obermayer.
\newblock Learning graph quantization.
\newblock In E.~R. Hancock, R.~C. Wilson, T.~Windeatt, I.~Ulusoy, and
  F.~Escolano, editors, \emph{Structural, Syntactic, and Statistical Pattern
  Recognition}, volume 6218, pages 109--118. Springer Berlin, Heidelberg, 2010.
\newblock \doi{10.1007/978-3-642-14980-1_10}.

\bibitem[Jothi and Rani(2015)]{jothi2015hybrid}
R.~B.~G. Jothi and S.~M.~M. Rani.
\newblock Hybrid neural network for classification of graph structured data.
\newblock \emph{International Journal of Machine Learning and Cybernetics},
  6\penalty0 (3):\penalty0 465--474, 2015.
\newblock \doi{10.1007/s13042-014-0230-8}.

\bibitem[Livi and Rizzi(2013)]{gm_survey}
L.~Livi and A.~Rizzi.
\newblock The graph matching problem.
\newblock \emph{Pattern Analysis and Applications}, 16\penalty0 (3):\penalty0
  253--283, 2013.
\newblock ISSN 1433-7541.
\newblock \doi{10.1007/s10044-012-0284-8}.

\bibitem[Livi et~al.(2013{\natexlab{a}})Livi, Bianchi, Rizzi, and
  Sadeghian]{odse2_ijcnn_2013}
L.~Livi, F.~M. Bianchi, A.~Rizzi, and A.~Sadeghian.
\newblock Dissimilarity space embedding of labeled graphs by a clustering-based
  compression procedure.
\newblock In \emph{{Proceedings of the IEEE International Joint Conference on
  Neural Networks}}, pages 1646--1653, Dallas, USA, Aug. 2013{\natexlab{a}}.
\newblock ISBN 978-1-4673-6129-3.
\newblock \doi{10.1109/IJCNN.2013.6706937}.

\bibitem[Livi et~al.(2013{\natexlab{b}})Livi, {Del Vescovo}, and
  Rizzi]{seriation+gradis_lncs_2012}
L.~Livi, G.~{Del Vescovo}, and A.~Rizzi.
\newblock Combining graph seriation and substructures mining for graph
  recognition.
\newblock In P.~{Latorre Carmona}, J.~S. S{\'a}nchez, and A.~L.~N. Fred,
  editors, \emph{{Pattern Recognition - Applications and Methods}}, volume 204,
  pages 79--91. Springer, Berling, Germany, 2013{\natexlab{b}}.
\newblock \doi{10.1007/978-3-642-36530-0_7}.

\bibitem[Livi et~al.(2014{\natexlab{a}})Livi, {Del Vescovo}, Rizzi, and
  {Frattale Mascioli}]{spare_graph_2013}
L.~Livi, G.~{Del Vescovo}, A.~Rizzi, and F.~M. {Frattale Mascioli}.
\newblock Building pattern recognition applications with the {SPARE} library.
\newblock \emph{ArXiv preprint arXiv:1410.5263}, Oct. 2014{\natexlab{a}}.

\bibitem[Livi et~al.(2014{\natexlab{b}})Livi, Rizzi, and Sadeghian]{odse}
L.~Livi, A.~Rizzi, and A.~Sadeghian.
\newblock Optimized dissimilarity space embedding for labeled graphs.
\newblock \emph{Information Sciences}, 266:\penalty0 47--64,
  2014{\natexlab{b}}.
\newblock ISSN 0020-0255.
\newblock \doi{10.1016/j.ins.2014.01.005}.

\bibitem[Livi et~al.(2015)Livi, Sadeghian, and Pedrycz]{eocc}
L.~Livi, A.~Sadeghian, and W.~Pedrycz.
\newblock Entropic one-class classifiers.
\newblock \emph{IEEE Transactions on Neural Networks and Learning Systems},
  26\penalty0 (12):\penalty0 3187--3200, Dec. 2015.
\newblock ISSN 2162-237X.
\newblock \doi{10.1109/TNNLS.2015.2418332}.

\bibitem[Livi et~al.(2016)Livi, Giuliani, and Rizzi]{ecoli_graph}
L.~Livi, A.~Giuliani, and A.~Rizzi.
\newblock Toward a multilevel representation of protein molecules:
  {C}omparative approaches to the aggregation/folding propensity problem.
\newblock \emph{Information Sciences}, 326:\penalty0 134--145, 2016.
\newblock ISSN 0020-0255.
\newblock \doi{10.1016/j.ins.2015.07.043}.

\bibitem[Lozano and Escolano(2006)]{Lozano2006539}
M.~A. Lozano and F.~Escolano.
\newblock {Protein classification by matching and clustering surface graphs}.
\newblock \emph{Pattern Recognition}, 39\penalty0 (4):\penalty0 539--551, 2006.
\newblock ISSN 0031-3203.
\newblock \doi{10.1016/j.patcog.2005.10.008}.

\bibitem[Lozano and Escolano(2013)]{Lozano2013177}
M.~A. Lozano and F.~Escolano.
\newblock {Graph matching and clustering using kernel attributes}.
\newblock \emph{Neurocomputing}, 113:\penalty0 177--194, 2013.
\newblock ISSN 0925-2312.
\newblock \doi{10.1016/j.neucom.2013.01.015}.

\bibitem[Luqman et~al.(2013)Luqman, Ramel, Llad{\'o}S, and
  Brouard]{Luqman:2013:FMG:2381462.2381562}
M.~M. Luqman, J.-Y. Ramel, J.~Llad{\'o}S, and T.~Brouard.
\newblock {Fuzzy multilevel graph embedding}.
\newblock \emph{Pattern Recognition}, 46\penalty0 (2):\penalty0 551--565, Feb.
  2013.
\newblock ISSN 0031-3203.

\bibitem[Marfil et~al.(2009)Marfil, Escolano, and
  Bandera]{marfil_escolano__2009}
R.~Marfil, F.~Escolano, and A.~Bandera.
\newblock {Graph-Based Representations in Pattern Recognition and Computational
  Intelligence}.
\newblock In J.~Cabestany, F.~Sandoval, A.~Prieto, and J.~M. Corchado, editors,
  \emph{{Bio-Inspired Systems: Computational and Ambient Intelligence}}, volume
  5517, pages 399--406. Springer Berlin, Heidelberg, 2009.
\newblock ISBN 978-3-642-02477-1.
\newblock \doi{10.1007/978-3-642-02478-8_50}.

\bibitem[Noma et~al.(2012)Noma, Graciano, {Cesar Jr}, Consularo, and
  Bloch]{Noma20121159}
A.~Noma, A.~B.~V. Graciano, R.~M. {Cesar Jr}, L.~A. Consularo, and I.~Bloch.
\newblock {Interactive image segmentation by matching attributed relational
  graphs}.
\newblock \emph{Pattern Recognition}, 45\penalty0 (3):\penalty0 1159--1179,
  2012.
\newblock ISSN 0031-3203.
\newblock \doi{10.1016/j.patcog.2011.08.017}.

\bibitem[P\c{e}kalska and Duin(2005)]{pkekalska+duin2005}
E.~P\c{e}kalska and R.~P.~W. Duin.
\newblock \emph{{The Dissimilarity Representation for Pattern Recognition:
  Foundations and Applications}}.
\newblock World Scientific, Singapore, 2005.

\bibitem[Porfiri et~al.(2008)Porfiri, Stilwell, and
  Bollt]{porfiri2008synchronization}
M.~Porfiri, D.~J. Stilwell, and E.~M. Bollt.
\newblock Synchronization in random weighted directed networks.
\newblock \emph{IEEE Transactions on Circuits and Systems I: Regular Papers},
  55\penalty0 (10):\penalty0 3170--3177, May 2008.
\newblock \doi{10.1109/TCSI.2008.925357}.

\bibitem[Pr{\'i}ncipe(2010)]{principe2010}
J.~C. Pr{\'i}ncipe.
\newblock \emph{{Information Theoretic Learning: Renyi's Entropy and Kernel
  Perspectives}}.
\newblock Springer-Verlag, NY, USA, 2010.

\bibitem[Ren et~al.(2011)Ren, Wilson, and Hancock]{5648360}
P.~Ren, R.~C. Wilson, and E.~R. Hancock.
\newblock {Graph Characterization via {I}hara Coefficients}.
\newblock \emph{IEEE Transactions on Neural Networks}, 22\penalty0
  (2):\penalty0 233--245, Feb. 2011.
\newblock ISSN 1045-9227.
\newblock \doi{10.1109/TNN.2010.2091969}.

\bibitem[Riesen and Bunke(2008)]{riesen+bunke2008}
K.~Riesen and H.~Bunke.
\newblock {IAM} graph database repository for graph based pattern recognition
  and machine learning.
\newblock In N.~{da Vitoria Lobo}, T.~Kasparis, F.~Roli, J.~T. Kwok,
  M.~Georgiopoulos, G.~C. Anagnostopoulos, and M.~Loog, editors,
  \emph{Structural, Syntactic, and Statistical Pattern Recognition}. Springer
  Berlin Heidelberg, Orlando, FL, 2008.
\newblock ISBN 978-3-540-89688-3.
\newblock \doi{10.1007/978-3-540-89689-0_33}.

\bibitem[Riesen and Bunke(2009{\natexlab{a}})]{riesen+bunke2009}
K.~Riesen and H.~Bunke.
\newblock {Graph classification by means of Lipschitz embedding}.
\newblock \emph{IEEE Transactions on Systems, Man, and Cybernetics, Part B},
  39:\penalty0 1472--1483, Dec. 2009{\natexlab{a}}.
\newblock ISSN 1083-4419.
\newblock \doi{10.1109/TSMCB.2009.2019264}.

\bibitem[Riesen and
  Bunke(2009{\natexlab{b}})]{riesen09Reducingdimensionalitydissimilarityspaceembeddinggraphkernels}
K.~Riesen and H.~Bunke.
\newblock {Reducing the dimensionality of dissimilarity space embedding graph
  kernels}.
\newblock \emph{Engineering Applications of Artificial Intelligence},
  22:\penalty0 48--56, Feb. 2009{\natexlab{b}}.
\newblock ISSN 0952-1976.
\newblock \doi{10.1016/j.engappai.2008.04.006}.

\bibitem[Riesen and Bunke(2010)]{riesen+bunke2010}
K.~Riesen and H.~Bunke.
\newblock \emph{{Graph Classification and Clustering Based on Vector Space
  Embedding}}.
\newblock World Scientific, Singapore, 2010.

\bibitem[Rizzi et~al.(2002)Rizzi, Panella, and {Frattale Mascioli}]{rizzi2002}
A.~Rizzi, M.~Panella, and F.~M. {Frattale Mascioli}.
\newblock Adaptive resolution min-max classifiers.
\newblock \emph{IEEE Transactions on Neural Networks}, 13:\penalty0 402--414,
  Mar. 2002.
\newblock ISSN 1045-9227.
\newblock \doi{10.1109/72.991426}.

\bibitem[Rizzi et~al.(2013)Rizzi, Colabrese, and Baiocchi]{6583538}
A.~Rizzi, S.~Colabrese, and A.~Baiocchi.
\newblock Low complexity, high performance neuro-fuzzy system for internet
  traffic flows early classification.
\newblock In \emph{Proceedings of the International Wireless Communications and
  Mobile Computing Conference}, pages 77--82, Sardinia, Jul. 2013.
\newblock \doi{10.1109/IWCMC.2013.6583538}.

\bibitem[Robles-Kelly and
  Hancock(2007)]{robleskelly07Riemannianapproachtographembedding}
A.~Robles-Kelly and E.~R. Hancock.
\newblock {A {R}iemannian approach to graph embedding}.
\newblock \emph{Pattern Recognition}, 40\penalty0 (3):\penalty0 1042--1056,
  2007.
\newblock \doi{10.1016/j.patcog.2006.05.031}.

\bibitem[Schleif and Ti{\v{n}}o(2015)]{schleif2015indefinite}
F.-M. Schleif and P.~Ti{\v{n}}o.
\newblock Indefinite proximity learning: {A} review.
\newblock \emph{Neural Computation}, 27\penalty0 (10):\penalty0 2039--2096,
  2015.
\newblock \doi{10.1162/NECO_a_00770}.

\bibitem[Serratosa et~al.(2013)Serratosa, Cort{\'e}s, and
  Sol{\'e}-Ribalta]{Serratosa:2013:CRB:2435460.2435704}
F.~Serratosa, X.~Cort{\'e}s, and A.~Sol{\'e}-Ribalta.
\newblock Component retrieval based on a database of graphs for hand-written
  electronic-scheme digitalisation.
\newblock \emph{Expert Systems with Applications}, 40\penalty0 (7):\penalty0
  2493--2502, Jun. 2013.
\newblock ISSN 0957-4174.

\bibitem[Shervashidze et~al.(2011)Shervashidze, Schweitzer, {van Leeuwen},
  Mehlhorn, and Borgwardt]{Shervashidze__2011}
N.~Shervashidze, P.~Schweitzer, E.~J. {van Leeuwen}, K.~Mehlhorn, and K.~M.
  Borgwardt.
\newblock {Weisfeiler-{L}ehman Graph Kernels}.
\newblock \emph{Journal of Machine Learning Research}, 12:\penalty0 2539--2561,
  Sep. 2011.
\newblock ISSN 1532-4435.

\bibitem[Wilson et~al.(2014)Wilson, Hancock, P\c{e}kalska, and
  Duin]{wilson2014spherical}
R.~C. Wilson, E.~R. Hancock, E.~P\c{e}kalska, and R.~P.~W. Duin.
\newblock Spherical and hyperbolic {E}mbeddings of data.
\newblock \emph{IEEE Transactions on Pattern Analysis and Machine
  Intelligence}, 36\penalty0 (11):\penalty0 2255--2269, Nov. 2014.
\newblock ISSN 0162-8828.
\newblock \doi{10.1109/TPAMI.2014.2316836}.

\end{thebibliography}
\end{document}